\documentclass[11pt]{article}

\usepackage[final]{acl}
\usepackage{multirow}
\usepackage{times}
\usepackage{latexsym}
\usepackage{amsmath}
\usepackage{subcaption}
\usepackage{stfloats}
\usepackage{xcolor}
\usepackage{fontawesome5}
\usepackage{hyperref}
\usepackage{algorithm}
\usepackage{algpseudocode}
\usepackage[normalem]{ulem} 
\usepackage{tabularx}
\usepackage{makecell}
\usepackage{placeins}
\usepackage{float}
\usepackage{enumitem}
\usepackage{minted}

\definecolor{skyblue}{RGB}{208,226,255}

\newcommand{\framework}{STaD}

\newcommand{\github}[1]{%
  \href{#1}{\faGithub}%
}

\newcommand{\database}[1]{%
  \href{#1}{\faDatabase}%
}

\usepackage[most]{tcolorbox}

\usepackage{amsmath, amssymb}

\usepackage{etoolbox}

\definecolor{lightgray}{RGB}{240,240,240}
\definecolor{darkgray}{RGB}{100,100,100}
\definecolor{lightblue}{RGB}{220,235,250}
\definecolor{lightgreen}{RGB}{220,250,220}
\definecolor{promptblue}{RGB}{0,75,150}

\usepackage{tikz}
\usepackage{collcell}
\usepackage[table]{xcolor}
\usepackage{etoolbox}

\newtoggle{inTableHeader}%
\toggletrue{inTableHeader}%
\newcommand*{\StartTableHeader}{\global\toggletrue{inTableHeader}}%

\let\OldTabular\tabular%
\let\OldEndTabular\endtabular%
\renewenvironment{tabular}{\StartTableHeader\OldTabular}{\OldEndTabular\StartTableHeader}%

\newcommand*{\MinNumber}{0.0}%
\newcommand*{\MidNumber}{0.5}%
\newcommand*{\MaxNumber}{1.0}%

\newcommand{\ApplyGradient}[1]{%
  \iftoggle{inTableHeader}{#1}{%
    \ifdim #1 pt > \MidNumber pt
        \pgfmathsetmacro{\PercentColor}{max(min(100.0*(#1-\MidNumber)/(\MaxNumber-\MidNumber),100.0),0.0)}%
        \colorbox{green!\PercentColor!yellow}{#1}%
    \else
        \pgfmathsetmacro{\PercentColor}{max(min(100.0*(\MidNumber-#1)/(\MidNumber-\MinNumber),100.0),0.0)}%
        \colorbox{red!\PercentColor!yellow}{#1}%
    \fi
  }%
}

\newcolumntype{R}{>{\collectcell\ApplyGradient}c<{\endcollectcell}}

\renewcommand{\arraystretch}{0.9}  
\usepackage[T1]{fontenc}

\usepackage[utf8]{inputenc}

\usepackage{microtype}

\usepackage{inconsolata}

\usepackage{graphicx}

\title{\framework: Scaffolded Task Design for Identifying Compositional Skill Gaps in LLMs}

\author{Sungeun An, Swanand Ravindra Kadhe, Shailja Thakur  \\ {\bf Chad DeLuca}, {\bf Hima Patel} \\ \\
        IBM Research \\
        \texttt{\{sungeun.an, swanand.kadhe, shailja.thakur\}@ibm.com}\\
        \texttt{delucac@us.ibm.com} \quad \texttt{himapatel@in.ibm.com}}

\begin{document}
\maketitle
\begin{abstract}
Benchmarks are often used as a standard to understand LLM capabilities in different domains. However, aggregate benchmark scores provide limited insight into compositional skill gaps of LLMs and how to improve them. To make these weaknesses visible, we propose \textbf{S}caffolded \textbf{Ta}sk \textbf{D}esign (\framework) framework. \framework~  generates controlled variations of benchmark tasks based on the concept of \textit{scaffolding}, which introduces structured, incremental support in a step-by-step manner. Rather than inspecting failures individually, this approach enables systematic and scalable probing of model behavior by identifying the specific reasoning skill compositions they lack. Treating the LLM as a black box, our experiments on six models of varying sizes reveal multiple failure points in three reasoning benchmarks and highlight each model's unique and distinct skill gaps.
\end{abstract}

Source Code: \github{https://github.com/ibm-granite/scaffolded-task-design}\quad Datasets: \database{https://huggingface.co/datasets/ibm-research/STaD}

\section{Introduction}

\begin{figure*}[!h]
    \centering
    \includegraphics[width=0.9\textwidth]{figs/scaffolded_competence.png}
    \caption{Scaffolded tasks for identifying compositional skill gap \textit{s2} (\textit{Subtraction}: Eggs remaining for sale).}
    \label{fig:scaffolding_example}
\end{figure*}

Benchmarks are standardized tests designed to evaluate the strengths and limitations of large language models (LLMs) across domains. However, most existing evaluations often treat complex tasks as monolithic \cite{ribeiro2020beyond, huang2023towards}. Like a single math test score, aggregate benchmark scores may suggest that a model is "bad at math" without explaining why. In particular, two models may achieve the same overall accuracy on the same benchmarks yet have entirely different skill gaps: one may lack a specific reasoning ability, another may struggle when multiple skills must be combined, and a third may have the necessary knowledge but fail to situate the knowledge in the context of the task.

Addressing such skill gaps requires examining a model's intermediate reasoning processes. One line of prior work has aimed to make model reasoning more explicit through Chain-of-Thought prompting, which elicits intermediate reasoning steps \cite{wei2022chain}. While this can surface intermediate traces of reasoning, these approaches are difficult to scale or analyze systematically \cite{huang2023towards}, and the generated explanations may not faithfully reflect the model’s actual internal reasoning process \cite{lanham2023measuring}. 

Another line of work focuses on task decomposition, which breaks complex tasks into simpler sub-tasks and makes intermediate reasoning steps more explicit and configurable \cite{khot2022decomposed, dua2022successive, zhou2022least, prasad2024adapt, jiang2023followbench, laban2025llms}. This provides a more structured representation of the reasoning process. However, existing decomposition-based methods are primarily used to improve task performance or to examine reasoning step by step, rather than to systematically diagnose failure modes.

From a cognitive science perspective, competence is not only what a learner or a model can do independently, but also what it can achieve with appropriate guidance \cite{vygotsky1978mind}. In this view, \textit{scaffolding} refers to targeted support that is gradually removed as proficiency develops \cite{van2010scaffolding}. We adapt this idea to LLM evaluation, using scaffolding not to improve performance, but as a diagnostic tool to reveal where and why models fail.

In this work, we propose \textbf{S}caffolded \textbf{Ta}sk \textbf{D}esign (\framework) framework to combine task decomposition and scaffolding to systematically identify the steps at which LLMs fail and the specific reasoning skill compositions they lack. In a nutshell, \framework~ breaks down benchmark tasks into sub-tasks representing individual reasoning steps; it then creates controlled variations of the original task by providing scaffolding at specific sub-tasks. By measuring performance across scaffolded variations, \framework~ can pinpoint exactly where the model struggles. Unlike isolated skill evaluation, we focus on compositional skill gaps that arise when multiple reasoning skills must be combined.

For example, consider the task in Figure~\ref{fig:scaffolding_example}, where a model must calculate Janet's daily earnings from selling eggs, which involves Addition, Subtraction, and Multiplication ($s_1$, $s_2$, and $s_3$). Suppose the model fails on the original task ($q$). We first scaffold $s_1$ (Addition: total eggs used daily), reducing the task to Subtraction and Multiplication. The model still fails, suggesting that the limitation is not confined to Addition. We then additionally scaffold $s_2$ (Subtraction: eggs remaining for sale), leaving only Multiplication. In this setting, the model succeeds and produces the correct answer of \$18. This reveals the model's compositional skill gap: it cannot independently perform Subtraction ($s_2$) in combination with Multiplication ($s_3$), even though it can handle Multiplication alone.

We evaluate \framework~ on three public reasoning benchmarks: ToT Arithmetic (Temporal Reasoning) \cite{fatemi2024test}, GSM8K (Grade School Math) \cite{cobbe2021gsm8k}, and Math-Hard (Difficult Math Problems) \cite{hendrycksmath2021}. Our results show that failures are multifaceted: models with similar overall performance can exhibit very different bottlenecks, some skills are present but fail in context, and compositional skill interaction frequently create challenges. Scaffolding further distinguishes tasks that can be reliably solved with targeted support from those that remain fundamentally intractable. \framework~ provides actionable insights, including: (1) Identifying skill combinations that cause failures; (2) Distinguishing weaknesses that can be mitigated with scaffolding from those that are fundamentally difficult; (3) Highlighting skill-level areas for further training or targeted interventions.

Our main contributions are threefold:
\begin{itemize}[noitemsep, topsep=0pt]
    \item \textbf{Concept}: We formalize \emph{scaffolded competence} for LLMs, capturing not only independent task performance but also the abilities a model can demonstrate with support. This distinguishes between tasks that become solvable with scaffolding and those that remain unsolved, revealing compositional skill gaps and failure modes.
    \item \textbf{Framework}: We introduce \textbf{S}caffolded \textbf{Ta}sk \textbf{D}esign (\framework) framework that generates scaffolded variations of benchmark tasks to systematically probe reasoning and compositional skill gaps in LLMs. We release the codebase and datasets, enabling application of \framework~ to existing benchmarks for constructing scaffolded task designs.
    \item \textbf{Empirical Study}: We apply our framework to three reasoning benchmarks. Our results reveal distinct skill gaps across models and demonstrate the value of detailed evaluation beyond aggregate scores.  
\end{itemize}

\section{Scaffolded Task Design (\framework)}
Given a benchmark, our goal is to construct scaffolded variations of its tasks that progressively introduce or remove support for intermediate reasoning steps. These scaffolded variations reveal latent competence and identify compositional skill gaps in the LLM. 

At a high level, \framework~ takes each benchmark question with its ground-truth answer and produces (1) a decomposition of the reasoning process into intermediate steps with corresponding partial answers, and (2) a set of scaffolded task variants that selectively provide support for these steps. We then evaluate models on both the original and scaffolded tasks to quantify \textit{scaffolded competence} and estimate the \textit{minimum level of scaffolding} required for successful reasoning.

Formally, we define a benchmark as a set of questions and corresponding answers, $Q = \{q_1, q_2, \dots, q_n\}$ and $A = \{a_1, a_2, \dots, a_n\}$. \framework~ operates using three functional LLM roles, which we describe in detail below.

\paragraph{Teacher model.}
The teacher model, $f_{\text{teach}}$, is responsible for generating scaffolded variations of each task including:
(i) decomposing $q$ into minimal reasoning units,
(ii) computing corresponding intermediate answers, and
(iii) constructing scaffolded task variants.
The teacher model can be implemented using different instantiations, including a single LLM, multiple models, or human-in-the-loop supervision.

\paragraph{Judge model.}
The judge model, $f_{\text{judge}}$, evaluates correctness for both intermediate and final predictions by comparing model outputs to the ground truth $a$. It returns a binary correctness score.

\paragraph{Target model.}
The target model, $f_{\text{target}}$, is the model we aim to diagnose. 
It is evaluated on both the original problem $q$ and the generated scaffolded variations $q_{\text{scaf}}$.

\subsection{Generating Test Cases}
\label{sec:generating_test_cases}

In this section, we describe how \framework~ generates test cases from a given benchmark. The process consists of four stages: (1) task decomposition, where each question is broken into a sequence of reasoning steps; (2) scaffolded task generation, where controlled variations are constructed; (3) verification, which filters inconsistent or unreliable scaffolded instances; and (4) skill mapping, which abstracts sub-tasks into higher-level reasoning skills to enable skill-level analysis.

\paragraph{Task Decomposition.}

\framework~ uses the teacher model $f_{\text{teach}}$ to decompose the original task $q$ into a sequence of $K$ sub-tasks $q \rightarrow S = \{s_1, s_2, \dots, s_K\}$, where each $s_k$ corresponds to a single logical or computational operation, the union $\bigcup_{k=1}^K s_k$ covers all reasoning steps required by $q$, and $s_K$ yields the final answer. This formulation is adapted from prior work on structured task decomposition \cite{laban2025llms}.

Given the decomposition $S$, the teacher model computes an intermediate answer for each sub-task, $sa_k = f_{\text{teach}}(s_k)$, resulting in $SA = \{sa_1, \dots, sa_K\}$. A binary consistency function $\mathrm{Cons}(sa_K, a)$ is defined, where $a$ denotes the ground-truth answer, returning 1 if the two values are identical and 0 otherwise.

\paragraph{Scaffolded Variations.}
\framework~ generates scaffolded variations of each original task $q$ by progressively revealing intermediate answers computed by the teacher model $f_{\text{teach}}$. Given $K$ sub-tasks, this yields $K-1$ scaffolded variations, each revealing solutions up to $s_{K-1}$ ($s_K$ is excluded as it yields the final answer). For each $j \in \{1, \dots, K-1\}$, the $j$-th scaffolded variation is constructed by injecting the first $j$ solved sub-tasks into the original task:
$q_{\text{scaf}}^{(j)} = g\!\left(q,\; \{(s_1,sa_1), \dots, (s_j,sa_j)\}\right)$,
where $g(\cdot)$ is a rewriting function that preserves the original task structure while explicitly providing the solutions to the first $j$ reasoning steps. Each $q_{\text{scaf}}^{(j)}$ is presented as a single instruction in which the first $j$ steps have already been completed, leaving the remaining steps to be solved by the model.

\paragraph{Verification.}\label{verification}
\framework~ verifies the consistency of all generated scaffolded variations. For each variation $q_{\text{scaf}}^{(j)}$, let ${p}_{\text{teach}}^{(j)} = f_{\text{teach}}(q_{\text{scaf}}^{(j)})$ denote the teacher model prediction. A binary consistency score is defined as $c_j = \mathrm{Cons}({p}_{\text{teach}}^{(j)}, a)$, where $c_j = 1$ if the prediction matches the ground-truth answer $a$, and $0$ otherwise. A scaffolded instance $q_{\text{scaf}}$ is retained only if the teacher model correctly solves all scaffolded variations, i.e., $c_j = 1$ for all $j \in \{1, \dots, K-1\}$. In other words, the instance is kept only when the teacher model can recover the correct final answer under every level of scaffolding. This requirement ensures the reliability of the decomposition. If any intermediate answer is incorrect, the resulting variation will contain incorrect information in its text. In such cases, the teacher model is unlikely to produce the correct final answer across all variations. These instances are therefore filtered out during verification.

\paragraph{Skill Mapping.}\label{skill_mapping}
To enable evaluation at the skill-level rather than individual sub-tasks, \framework~ maps sub-tasks into higher-level skills that reflect shared underlying cognitive or reasoning capabilities. \framework~ first embeds each task into a semantic vector space and clusters tasks using Agglomerative Clustering  \cite{pedregosa2011scikit}. From each cluster, it then selects a representative task whose embedding is closest to the cluster centroid. These representative tasks are used by $f_{\text{teach}}$ to induce a set of higher-level skills:
$\mathcal{H} = \{h_1, h_2, \dots, h_M\}$. Each sub-task $s_k$ is then assigned to one skill in $\mathcal{H}$. This mapping abstracts away from individual sub-tasks and enables evaluation at the higher-level skill.

\subsection{Target Model Evaluation}
Once the test cases are generated, the target model $f_{\text{target}}$ is evaluated on (i) the original task $q$, and (ii) a set of scaffolded variations $q_{\text{scaf}}$. These test cases reveal which reasoning steps the model can handle independently and which require external support, enabling fine-grained analysis of failure points in multi-step reasoning.

\paragraph{Original Task Evaluation.}
The target model $f_{\text{target}}$ generates predictions for the original benchmark questions, denoted as ${P}_{\text{orig}} = \{{p}_1, \dots, {p}_n\}$. To evaluate correctness, \framework~ applies the consistency function $\mathrm{Cons}$ to each prediction against its corresponding ground-truth answer. This yields a vector of binary scores over the dataset:
$\mathrm{score}_{\text{orig}} = \big[\,\mathrm{Cons}({p}_i, a_i)\,\big]_{i=1}^n$.

\paragraph{Scaffolded Task Evaluation.}\label{test_evaluation}
For each question $q_i$ that the target model initially answers incorrectly (i.e., $\mathrm{score}_{\text{orig}}^i = 0$), \framework~ evaluates the model on the corresponding scaffolded variations $q_{\text{scaf}}^{(1)}, \dots, q_{\text{scaf}}^{(K-1)}$ to measure how performance changes as additional intermediate steps are revealed. For each variation, correctness is assessed using the consistency function $\mathrm{Cons}$, yielding a sequence of binary outcomes:
$\mathrm{score}_{\text{scaf}} = \big[\,\mathrm{Cons}(\hat{p}_i^{(j)}, a_i)\,\big]_{j=1}^{K-1}$.

\subsection{Minimum Scaffolding Level $k$}\label{sec:k}
We define the \emph{minimum scaffolding level} $k$ for question $q$ as the smallest number of cumulative scaffolding required for the model to answer a task correctly:
\[
\begin{array}{rcl}
k &=& \min \Big\{ j \mid \mathrm{Cons}(\hat{p}^{(j)}, a) = 1 \\ 
& & \:\:\qquad \wedge\: \mathrm{Cons}(\hat{p}^{(l)}, a) = 0, l < j \Big\}.
\end{array}
\]

If no scaffolded variation yields a correct prediction, we set $k = -1$. The minimum scaffolding level provides a quantitative measure of model competence: (i) Tasks are categorized as $k=0$ if solvable independently, (ii) $k>0$ if solvable with scaffolding, and (iii) $k=-1$ if unsolvable even with full scaffolding. Intuitively, higher values of $k$ indicate that more guidance is required, and therefore correspond to greater task difficulty.

\subsection{Individual Skill Testing as a Baseline}
As a baseline, individual skills are evaluated by testing each sub-task in isolation. This corresponds to a standard evaluation setup that measures model performance at the level of individual skills, without considering interactions across multiple reasoning steps. Each sub-task is associated with a higher-level skill (as described in Section~\ref{skill_mapping}). For each skill, accuracy is computed as the proportion of correctly solved sub-tasks belonging to that skill. This measure reflects how well the model performs on isolated skills.

\section{Experiment Setup}
We instantiate \framework~ and describe its empirical implementation.

\paragraph{Datasets.}
We evaluate \framework~ on three benchmark datasets: \textbf{ToT (Test of Time) Arithmetic} \cite{fatemi2024test}, \textbf{GSM8K (Grade School Math 8K)} \cite{cobbe2021gsm8k}, and \textbf{Math-Hard} \cite{hendrycksmath2021}. We focus on these benchmarks as they are popular for assessing reasoning in small LMs, span varying difficulty levels, and together cover a broad set of skills. While our study focuses on these datasets, \framework~ can be applied to any benchmark that involves multi-step reasoning with well-defined intermediate answers.

\paragraph{Models.}
We instantiate \framework~ with GPT-OSS-120B \cite{agarwal2025gpt} as the \emph{teacher model} for test case generation. For evaluation, we use two \emph{judge models}: Llama-3.3-70B \cite{grattafiori2024llama3herdmodels} for verification (Section \ref{verification}), and Mistral-Large \cite{Mistral} for evaluation (Section \ref{test_evaluation}). To analyze skill gaps, we evaluate a set of \emph{target models} of varying sizes and families: Qwen2.5-3B-Instruct and Qwen2.5-7B-Instruct \cite{qwen2025qwen25technicalreport}, Granite-3.3-2B-Instruct and Granite-3.3-8B-Instruct \cite{granite2024granite}, and Llama-3.2-3B-Instruct and Llama-3.1-8B-Instruct \cite{grattafiori2024llama3herdmodels}. For test case generation, we use a temperature of 0.2 with the teacher model to introduce mild generation diversity. All evaluations with the judge and target models use greedy decoding, ensuring deterministic and consistent results across runs.

\paragraph{Dataset Construction.}
As described in Section~\ref{sec:generating_test_cases}, \framework~ filters scaffolded instances based on verification consistency. Specifically, an instance (original problem $q$) is retained only if the teacher model produces the correct answer for all of its scaffolded variations $\{q_{\text{scaf}}^{(j)}\}_{j=1}^{K-1}$. This filtering criterion ensures that only instances with fully consistent and reliable decomposition structures are included in the final dataset. Table~\ref{tab:scaffolded_datasets} summarizes the resulting dataset statistics, including the original and filtered (verified) dataset sizes and the average number of sub-tasks per example (min/max). The final filtered dataset size is indicated by the right arrow.

\begin{table}[h]
\centering
\caption{Dataset statistics before and after filtering, and average number of sub-tasks (min, max).}
\begin{tabular}{lccc}
\hline
\textbf{Dataset}  & \textbf{Size} & \textbf{Avg Sub-Tasks }  \\
\hline
ToT  & 1.85K$\rightarrow$1.45K & 4.59 (2, 6) \\
GSM8K  & 1.31K$\rightarrow$1.17K  & 3.92 (2, 7)  \\
Math-Hard  & 1.32K$\rightarrow$773 & 5.48 (2, 6) \\
\hline
\end{tabular}
\label{tab:scaffolded_datasets}
\end{table}

\begin{figure*}[t]
  \centering
  \includegraphics[width=\textwidth]{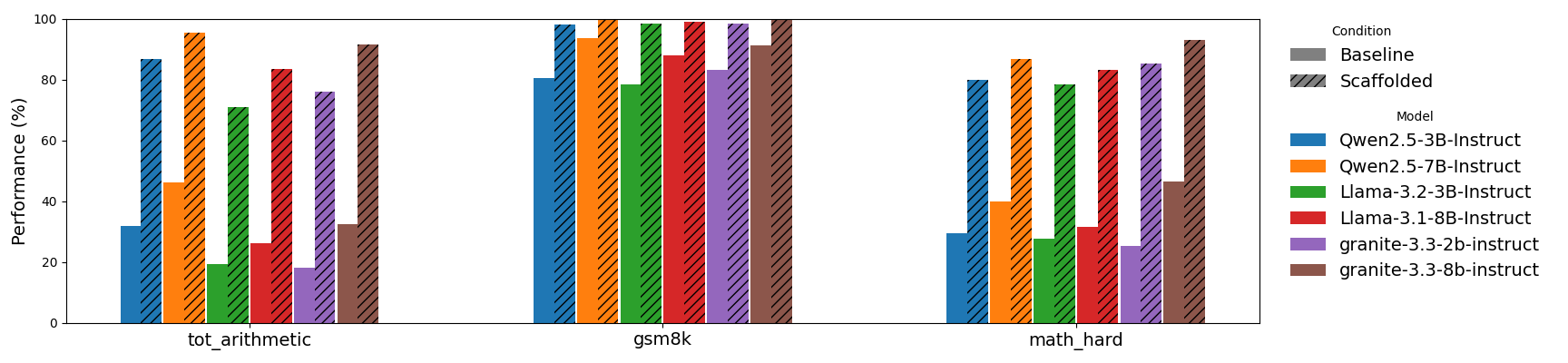}
  \caption{Original vs.\ scaffolded performance across benchmarks.}
  \label{fig:benchmark_evaluation}
\end{figure*}

\paragraph{Decomposition Quality.}
We conduct a quantitative evaluation of decomposition quality in terms of redundancy and step coverage. Redundancy measures the extent to which sub-tasks repeat similar information, while step coverage assesses how comprehensively the decomposition captures the reasoning required to solve the original instance. The results show consistently low redundancy (4.21--8.29\%) and high coverage (87.52--99.31\%), indicating that the decomposition yields diverse and complete intermediate reasoning steps (see Appendix~\ref{appendix:scaffold_quality}).

\paragraph{Skill Granularity.}
We selected the \textit{right} level of granularity in terms of the number of representative tasks and number of high-level skills via careful ablation experiments (see Appendix~\ref{appendix:cluster_skill}). For ToT Arithmetic, we selected 40 representative questions per cluster and identified 20 skills; for GSM8K, 40 questions generated 40 skills; and for Math-Hard, 40 questions generated 20 skills. The full skill lists are provided in Appendix~\ref{appendix:skill_list}.

\section{Results and Analysis}
We now present the results of applying \framework~ with generated scaffolding across benchmarks.

\subsection{Benchmark Evaluation}
Figure~\ref{fig:benchmark_evaluation} shows model performance on original tasks (baseline) and under scaffolding across three benchmarks. A task is considered successful if any scaffolded variant yields the correct answer. In the original setting, GSM8K shows strong performance across models (78.4\%--93.62\%), while ToT and Math-Hard are significantly lower (18.16\%--46.12\% and 25.36\%--46.44\%, respectively). Under scaffolding, performance increases across all benchmarks. Importantly, this does not reflect improved model capability, but instead shows whether failures can be resolved when intermediate reasoning is provided. When a task becomes solvable under scaffolding, it indicates that the original failure was due to missing intermediate skills, revealing a reasoning gap. Gains are larger on ToT and Math-Hard, while GSM8K shows smaller improvements due to its already high baseline.

We conducted ablation experiments confirming the scaffolding effect is distinct from information leakage: replacing injected values with non-informative placeholder text drops accuracy from 100\% to $\sim$12\% across three models (Appendix~\ref{app:ablation_leakage}).

\subsection{Individual Skill Gaps}
Table~\ref{tab:method1_skill_accuracy_per_dataset} describes independent skill accuracy. Higher accuracy indicates stronger performance, while lower accuracy indicates skill weaknesses. Individual skill testing reveals model strengths and weaknesses in an aggregated view. However, skills are evaluated in isolation without interactions or situational dependencies, which may provide an overly optimistic view of actual task performance.

In ToT Arithmetic, models perform well on \textit{Structured JSON output} (0.38--0.70), \textit{Arithmetic on durations} (0.42--0.71), and \textit{Multi-format date conversion} (0.42--0.65), but struggle on \textit{Chronological ordering of events} (0.02--0.14), \textit{Overlap and intersection of time} (0.11--0.22), and \textit{Calendar arithmetic} (0.16--0.23). In GSM8K, models are generally strong across most skills, with few exceptions such as \textit{Extracting quantitative data from text} (0.11--0.29) and \textit{Formulating and solving linear equations} (0.42--0.50). In Math-Hard, skill performance is more inconsistent across models, with tasks such as \textit{Translating problems into algebra} remaining challenging (0.23--0.34).

\begin{table*}[ht]
\setlength{\abovecaptionskip}{3pt}
\centering
\small
\setlength{\tabcolsep}{5pt}
\caption{Individual skill accuracy across models and datasets (top 10 most frequent skills). The two lowest-performing skills per dataset are highlighted in bold.}

\label{tab:method1_skill_accuracy_per_dataset}
\begin{tabular}{ll|cccccc}
\hline
&\textbf{Skill} 
& \textbf{Qwen3B} & \textbf{Qwen7B} 
& \textbf{Llama3B} & \textbf{Llama8B} 
& \textbf{Granite2B} & \textbf{Granite8B} \\
\hline
\multirow{10}{*}{\rotatebox{90}{\textit{ToT Arithmetic}}}
& Structured JSON output & 0.55 & 0.70 & 0.38 & 0.42 & 0.53 & 0.66 \\
&Arithmetic on durations & 0.54 & 0.71 & 0.42 & 0.50 & 0.55 & 0.69 \\
&Unit conversion for time spans & 0.51 & 0.70 & 0.33 & 0.42 & 0.37 & 0.60 \\
&Calendar arithmetic & 0.18 & 0.22 & 0.19 & \textbf{0.18} & 0.16 & 0.23 \\
&Natural-language date/time parsing & 0.35 & 0.27 & 0.23 & 0.26 & 0.28 & 0.27 \\
&Discrete slot counting & 0.19 & 0.29 & 0.15 & 0.21 & 0.16 & 0.25 \\
&Overlap and intersection of time & \textbf{0.11} & \textbf{0.18} & \textbf{0.13} & 0.20 & \textbf{0.11} & \textbf{0.22} \\
&Multi-format date conversion & 0.62 & 0.65 & 0.42 & 0.52 & 0.55 & 0.64 \\
&Day-of-week determination & 0.44 & 0.37 & 0.27 & 0.36 & 0.28 & 0.43 \\
&Chronological ordering of events & \textbf{0.02} & \textbf{0.04} & \textbf{0.08} & \textbf{0.14} & \textbf{0.08} & \textbf{0.10} \\

\hline
\multirow{10}{*}{\rotatebox{90}{\textit{GSM8K}}}
& Sequential quantity tracking & 0.70 & 0.77 & 0.74 & 0.70 & 0.66 & 0.77 \\
& Using unit rates to compute totals & 0.82 & 0.82 & 0.86 & 0.87 & 0.75 & 0.83 \\
&Extracting quantitative data from text & \textbf{0.25} & \textbf{0.19} & \textbf{0.29} & \textbf{0.20} & \textbf{0.11} & \textbf{0.19} \\
&Summing contributions from multi sources & 0.67 & 0.78 & 0.77 & 0.80 & 0.69 & 0.89 \\
&Formulating and solving linear equations & \textbf{0.49} & \textbf{0.50} & \textbf{0.42} & \textbf{0.48} & 0.51 & \textbf{0.42} \\
&Multi-digit multiplication and division & 0.90 & 0.85 & 0.84 & 0.94 & 0.73 & 0.83 \\
&Proportional reasoning & 0.77 & 0.76 & 0.77 & 0.85 & 0.57 & 0.79 \\
&Calculating a percentage of a quantity & 0.85 & 0.89 & 0.93 & 0.96 & 0.67 & 0.93 \\
&Performing operations with fractions & 0.70 & 0.80 & 0.73 & 0.69 & 0.60 & 0.58 \\
&Multiplicative scaling for unknowns & 0.81 & 0.86 & 0.76 & 0.73 & 0.74 & 0.75 \\

\hline
\multirow{10}{*}{\rotatebox{90}{\textit{Math-Hard}}}
&Multi-step integer/fraction arithmetic & 0.45 & 0.51 & 0.55 & 0.60 & 0.56 & 0.63 \\
&Translating problems into algebra & \textbf{0.24} & \textbf{0.23} & \textbf{0.31} & \textbf{0.31} & \textbf{0.34} & \textbf{0.34} \\
&Solving linear equations and systems & 0.29 & 0.31 & 0.41 & 0.45 & 0.46 & 0.47 \\
&Combinatorial counting & 0.36 & 0.46 & 0.44 & 0.42 & 0.43 & 0.47 \\
&Quadratic equation techniques & 0.30 & 0.40 & 0.44 & 0.51 & 0.50 & 0.42 \\
&Modular arithmetic and congruence & 0.34 & 0.44 & 0.40 & \textbf{0.35} & \textbf{0.39} & 0.54 \\
&GCD/LCM via prime factorization & 0.38 & 0.41 & 0.49 & 0.51 & 0.49 & 0.53 \\
&Probability via counting & 0.28 & 0.46 & \textbf{0.39} & 0.39 & 0.44 & 0.49 \\
&Domain and range of functions & \textbf{0.20} & \textbf{0.20} & 0.55 & 0.49 & 0.43 & 0.50 \\
&Arithmetic and geometric series analysis & 0.36 & 0.28 & 0.43 & 0.41 & 0.52 & \textbf{0.35} \\

\hline
\end{tabular}

\end{table*}

\subsection{Minimum Scaffolding: Bottleneck} 
We define the minimum scaffolding set as a \textit{bottleneck set}: the smallest set of reasoning steps that must be externally supported for the model to reach a correct solution. The bottleneck identifies the earliest point at which reasoning fails, and thus reflects a \textit{compositional skill gap}: a failure mode in which a target skill breaks down when it must be coordinated with other interacting skills, even though that same skill may succeed in isolation. This answers: \emph{Where does the LLM’s reasoning first fail?}

\begin{figure*}[!ht]
    \centering
    \includegraphics[width=\textwidth]{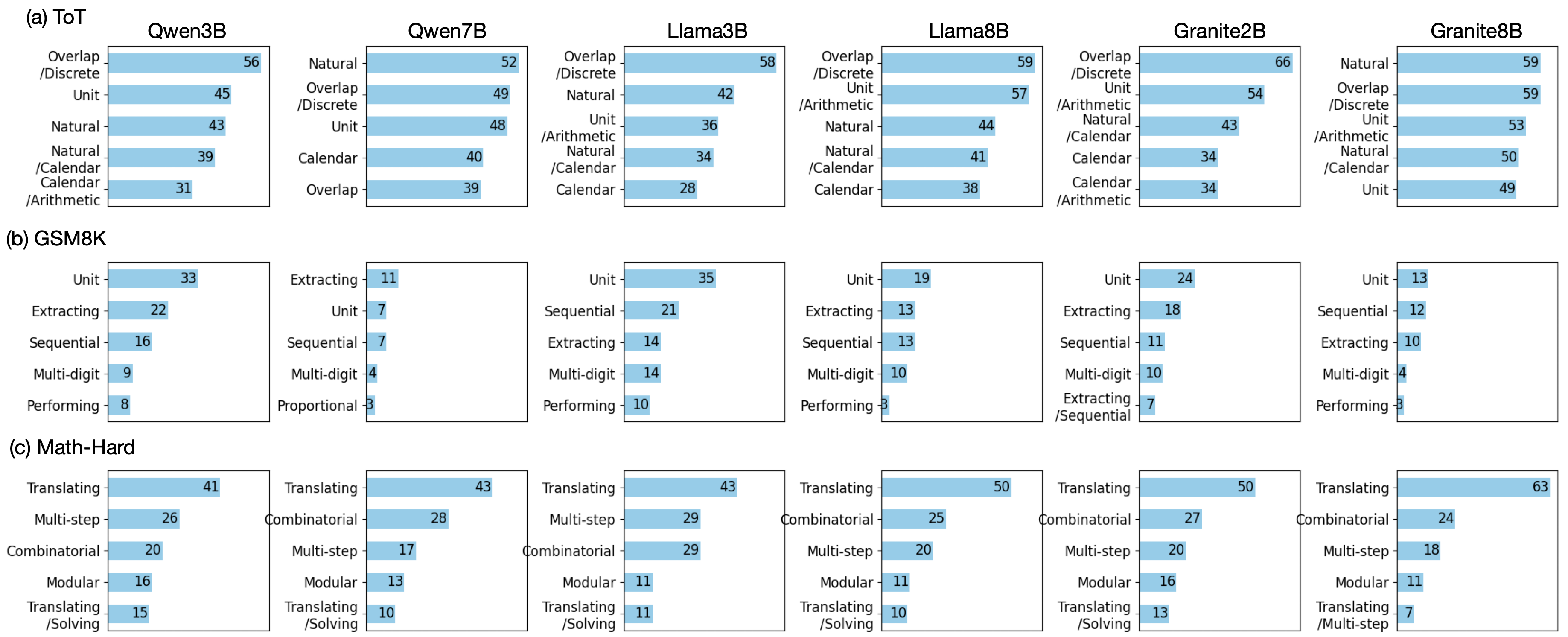}
    \caption{Frequency of compositional-skill bottlenecks in (a) ToT Arithmetic, (b) GSM8K, and (c) Math-Hard. For each model, the top 5 bottlenecks are displayed. Numbers indicate the count of scaffolded cases. Only the first word of each skill is shown.}
    \label{fig:skill_bottleneck}
    
\end{figure*}

For analysis, we normalize the data by collapsing duplicated skills (e.g., "Arithmetic on durations, Arithmetic on durations" to "Arithmetic on durations") since repeated scaffolding of the same skill reflects persistent difficulty with that skill rather than additional reasoning structure. Figure \ref{fig:skill_bottleneck} shows the frequency of the top 5 bottlenecks. Full results are provided in Table~\ref{tab:skill_bottlenecks_top10}.

\paragraph{ToT Arithmetic} \textit{Overlap time/Discrete slot} (49--66), \textit{Natural-language date/time parsing} (30--59), and \textit{Natural-language/Calendar arithmetic} (30--50) are the most frequent bottlenecks. \textit{Unit conversion/Arithmetic on durations} (17--57) varies widely across models, being a major bottleneck for Llama8B, Granite2B, and Granite8B but minor for Qwen7B.

\paragraph{GSM8K} Frequent bottlenecks include \textit{Using unit rates} (7--35), \textit{Extracting quantitative data} (10--22), and \textit{Sequential quantity tracking} (7--21). Bottleneck frequencies are relatively low overall, indicating no single skill strongly blocks progress.

\paragraph{Math-Hard} \textit{Translating word problems} (41--63) is the dominant bottleneck, followed by \textit{Multi-step reasoning} (17--29) and \textit{Combinatorial reasoning} (20--29). Multi-skill combinations (\textit{Translating/Solving}, \textit{Translating/Multi-step}) appear at moderate frequencies, while larger combinations ($size \geq 3$) are rare.

\begin{tcolorbox}[colback=skyblue, coltext=black, colframe=skyblue, width=\linewidth, boxrule=0.5mm, arc=1mm]
\textbf{Takeaways: } The most-frequent bottleneck analysis identifies which skills fail most often. LLM failures stem from early breakdowns in coordinating multiple skills. 
\end{tcolorbox}

\subsection{Mean Minimum Scaffolding per Reasoning Combination}
We analyze the mean minimum scaffolding for tasks that share the same underlying skill combination. Specifically, tasks are grouped by their original skill combinations, and for each task \framework~ computes the minimum scaffolding level ($k$) required for the model to produce a correct solution (Section~\ref{sec:k}). Intuitively, the minimum scaffolding level ($k$) represents the least amount of external guidance needed for success; larger values of $k$ therefore indicate greater difficulty. Table~\ref{tab:scaffolding} reports results for the most frequent skill combinations in ToT benchmark. Full results are provided in Table~\ref{tab:scaffolding_copy}.

The combination \textit{Overlapping time + Overlapping time + Discrete slots + Discrete slots + JSON} is consistently challenging across models. This is reflected in both high mean $k$ values and a substantial fraction of intractable tasks. For example, Granite2B fails on 24.61\% of these problems even with full scaffolding. Across models, mean $k$ ranges from 2.51 to 3.06, indicating that solving these tasks typically requires two to three levels of explicit guidance (out of four available scaffolding). In practice, this typically means providing 2-3 levels of guidance—for example, first guiding the model through \textit{Overlapping}, then through \textit{Discrete Meeting slots}—after which it can usually handle the remaining \textit{Discrete Meeting slots} and \textit{JSON} components on its own.

The combination \textit{Unit conversion + Arithmetic on durations  + Arithmetic on durations + Unit conversion + JSON} reveals model-dependent skill gaps. Qwen3B, Qwen7B, and Granite8B handle this combination with relatively little assistance, exhibiting mean $k$ values between 1.45 and 1.97 and near-zero intractable rates (0-1.63\%). However, Llama3B, Llama8B, and Granite2B struggle substantially, with mean $k$ values between 3.19 and 3.68 and intractable rates ranging from 27.86\% to 59.01\%. This divergence demonstrates that identical composite skills can be trivial for some models while remaining fundamentally difficult for others. A similar pattern is observed for the combination \textit{Unit + Midnight + Arithmetic + Unit + JSON}.

In GSM8K, most tasks are solved independently without any scaffolding: over 90\% of problems have $k=0$ across all models, and almost none remain intractable once scaffolding is introduced. For GSM8K and Math-Hard, however, we were unable to collect enough tasks sharing identical skill combinations to perform a comparable combination-level analysis. This reflects the higher structural diversity of these benchmarks, where problems vary widely in composition rather than recurring around a small set of shared skill combinations.

\begin{tcolorbox}[colback=skyblue, coltext=black, colframe=skyblue, width=\linewidth, boxrule=0.5mm, arc=1mm]
\textbf{Takeaways: } Mean minimum scaffolding reveals how difficult specific skill combinations are to resolve. Some combinations are universally hard, while others show model-specific gaps.
\end{tcolorbox}

\section{Discussion}

Existing probing methods, such as directly asking whether a model can perform a skill, abstract away the context in which a skill must be applied. In real tasks, competence is expressed through the coordinated deployment of multiple skills, situated within the specific demands of the task context. By preserving this context while selectively supporting intermediate reasoning stages, scaffolding enables direct evaluation of situated competence---the model's ability to apply a given skill appropriately within its original task setting.

\paragraph{Comparable performance does not imply identical skill gaps.} While overall task scores may be similar, models often exhibit very different skill bottleneck profiles. For example, Qwen3B and Granite8B achieve similar ToT performance (31.84 vs. 32.46), but certain reasoning combinations (e.g., \textit{Overlapping time + Overlapping time + Discrete slots + Discrete slots + JSON}) was challenging in different levels as reflected by mean $k$ and a substantial fraction of intractable tasks for Qwen3B (23.07\%) but less for Granite8B (7.69\%). In another case, Llama3B and Granite2B achieve similar ToT performance (19.2 vs. 18.16), but Llama3B tends to require more support on isolated skills (e.g., \textit{Natural-language date/time parsing} and \textit{Calendar arithmetic}) whereas Granite2B's bottlenecks appear when multiple skills are combined (e.g., \textit{Unit conversion / Arithmetic on durations},\textit{Relative-date reasoning / Calendar arithmetic / Multi-format date conversion}). Comparable trends appear in GSM8K. Qwen3B and Granite2B achieve similar overall accuracy (80.36 vs. 83.08), yet Qwen3B exhibits high individual skill averages ($0.74 \pm 0.17$) while Granite2B shows lower individual skill competence ($0.62 \pm 0.17$), particularly struggling with tasks like \textit{Extracting quantitative data} from text that Qwen3B handles reliably. 

\begin{table*}[!ht]
\setlength{\abovecaptionskip}{5pt}
\centering
\small
\setlength{\tabcolsep}{8pt}
\caption{Mean minimum scaffolding level ($k$) per skill combination. $k=0$: solved independently; \%\,$k>0$: solvable with scaffolding; \% Intractable: unsolvable even with full scaffolding. Higher $k$ indicates greater difficulty.}
\label{tab:scaffolding}
\renewcommand{\arraystretch}{1.05}
\begin{tabular}{lcccccc}
\hline
\textbf{Combination (ToT)} & \textbf{\#Tasks} & \textbf{Model} & \textbf{Mean $k$} & \textbf{\% $k=0$} & \textbf{\% $k>0$} & \textbf{\% Intractable} \\
\hline
\multirow{6}{*}{\shortstack[l]{\textit{Overlapping} + \textit{Overlapping} + \\
\textit{Discrete} + \textit{Discrete} + \\
\textit{JSON}}}
& \multirow{6}{*}{65} & Qwen3B    & 3.02 & 6.15 & 70.76 & 23.07 \\
&                     & Qwen7B    & 2.51 & 6.15 & 89.23 & 4.61 \\
&                     & Llama3B   & 3.06 & 3.07 & 75.38 & 21.53 \\
&                     & Llama8B   & 2.69 & 10.76 & 81.53 & 7.69 \\
&                     & Granite2B & 2.93 & 1.53 & 73.84 & 24.61 \\
&                     & Granite8B & 2.79 & 10.76 & 81.53 & 7.69 \\
\hline
\multirow{6}{*}{\shortstack[l]{\textit{Unit} + \textit{Arithmetic} + \textit{Arithmetic} + \\
\textit{Unit} + \textit{JSON}}}
& \multirow{6}{*}{61} & Qwen3B    & 1.90 & 45.90 & 52.45 & 1.63 \\
&                     & Qwen7B    & 1.45 & 67.21 & 32.78 & 0.0 \\
&                     & Llama3B   & 3.68 & 9.83 & 31.14 & 59.01 \\
&                     & Llama8B   & 3.44 & 13.11 & 59.01 & 27.86 \\
&                     & Granite2B & 3.19 & 9.83 & 59.01 & 31.14 \\
&                     & Granite8B & 1.97 & 21.31 & 78.68 & 0.0 \\
\hline
\multirow{6}{*}{\shortstack[l]{\textit{Unit} + \textit{Midnight} + \\
\textit{Arithmetic} + \textit{Unit} + \textit{JSON}}}
& \multirow{6}{*}{41} & Qwen3B    & 2.32 & 4.87 & 90.24 & 4.87 \\
&                     & Qwen7B    & 2.18 & 34.14 & 65.85 & 0.00 \\
&                     & Llama3B   & 3.85 & 0.0 & 68.29 & 31.70 \\
&                     & Llama8B   & 3.60 & 0.0 & 80.48 & 19.51 \\
&                     & Granite2B & 3.82 & 2.43 & 68.29 & 29.26 \\
&                     & Granite8B & 2.48 & 9.75 & 85.36 & 4.87 \\
\hline
\end{tabular}
\end{table*}

\paragraph{Skill Present but Fail in Context.} Performance on isolated skill probes systematically overestimates a model's actual reasoning competence when those same skills must be executed in combination. Several abilities that appear relatively well mastered in isolation, such as \textit{Arithmetic on durations}(0.42-0.71), \textit{Unit conversion for time spans} (0.33-0.70), and \textit{Natural-language date/time parsing} (0.23-0.35), \textit{Combinatorial counting} (0.36-0.47) do not emerge as dominant failure skills under individual skill testing. However, scaffolding analysis reveals that these skills become among the most frequent bottlenecks when embedded within tasks that require sequential reasoning steps, both as single-skill and compositional-skill bottlenecks. More broadly, these results suggest that model reasoning weaknesses are not solely a function of missing skills, but of unstable skill integration. Models may possess the necessary primitives yet lack mechanisms for sequencing, maintaining intermediate representations, or reconciling constraints across skills.

\paragraph{Failures often arise from skill interactions.} Compositional skill bottlenecks demonstrate that improving only the upstream skills is insufficient. For example, scaffolding both \textit{Overlap} and \textit{Discrete} skills together nearly doubles the success rate compared to scaffolding \textit{Overlap} alone (44\% vs.\ 22\% for Qwen7B on ToT Arithmetic). In particular, tasks that require \textit{Overlap} + \textit{Overlap} + \textit{Discrete Meeting slots} + \textit{Discrete Meeting slots} + \textit{JSON} indicate that, on average, models require support beyond pure \textit{Overlap} reasoning including \textit{Discrete Meeting slots} skills. This suggests that, in compositional tasks, addressing only the upstream skills is unlikely to improve performance, and that failures often stem not from deficiencies in individual skills, but from their interactions—a model may handle each skill independently yet fail when they must be coordinated.

\paragraph{Scaffolding Reveals Learnable vs. Intractable Tasks.} Scaffolding analysis highlights which tasks can be solved when upstream skills are supported and which remain intractable despite full support. Successful scaffolded cases demonstrate that these tasks are learnable: when the model receives guidance for the necessary intermediate skills, it is able to reach the correct solution. Tasks that remain unsolved, such as those in ToT arithmetic and Math-Hard, point to deeper challenges beyond isolated skill gaps. One contributing factor is that scaffolding was provided only up to the final step, so failures may reflect difficulty in synthesizing intermediate results or completing the last step. Other failures may stem from challenges in understanding the original question or in planning a decomposition strategy, suggesting that intractable tasks require more than just learning individual skills. They demand higher-level reasoning to integrate multiple components.  

\paragraph{Complementary Approaches for Diagnosing Model Skill Gaps.} 
Individual skill testing and scaffolding serve complementary roles: the former identifies what a model can do in isolation, while the latter reveals how those capabilities interact during end-to-end reasoning. Decomposition is good for isolating specific skills or reasoning steps that are absent, offering a precise map of the task's underlying components and making localized skill deficits visible. Scaffolding, by contrast, situates these skills within the broader problem context, revealing not only whether a model can apply a skill in practice but also how multiple skills interact. Beyond assessing competence, scaffolding exposes which tasks remain fundamentally intractable and which can be resolved with targeted support, highlighting both the limits of model reasoning and the actionable pathways for improvement.

\paragraph{Actionable Implications.} These findings suggest concrete directions: (1) targeted synthetic data generation at the skill-combination level rather than individual skills; (2) evaluation protocols that report both isolated skill accuracy and compositional bottlenecks; and (3) training strategies that explicitly target skill coordination, not just skill acquisition.

\section{Conclusions}

Through applying our framework to three benchmarks and six models, we discovered that not all reasoning failures in LLMs are created equal. Some models stumble on the basics: they cannot perform individual reasoning steps reliably. Other models, while capable of these individual skills, struggle when multiple skills must work together, revealing that composing knowledge is a challenge. Finally, some models appear competent on every step in isolation but fail when they need to apply that knowledge in context. By combining task decomposition with scaffolding, we were able to trace these different failure modes systematically, offering a lens to see beyond aggregate benchmark scores. These findings provide actionable insights: creating targeted synthetic data at the right level of abstraction, refining training strategies, and designing evaluation frameworks that emphasize both individual skills and their composition. We release all code and datasets to support reproducibility and further research.

\newpage
\section*{Limitations}

\paragraph{Teacher model dependence.} Our approach relies on a teacher model to generate scaffolded task variations, so decomposition quality depends on teacher capability. We note that multiple valid decompositions exist, and different strategies may yield alternative sub-task sequences and insights. However, \framework~is agnostic to the choice of teacher and is fully compatible with stronger models, human-in-the-loop teachers, or ensemble approaches. To empirically assess robustness, we validate cross-teacher agreement using two independent teachers: on average, 3.7 out of 5 top bottleneck categories overlap across teachers, with one model achieving perfect agreement (5/5). Remaining discrepancies involve closely related skills (e.g., \textit{Calendar} vs.\ \textit{Calendar/Arithmetic}), indicating minimal practical disagreement (Appendix~\ref{app:cross_teacher}).

\paragraph{Scope.} \framework~is designed for multi-step reasoning tasks where questions share similar reasoning structures, enabling aggregation across failures. It is less applicable to single-step or open-ended generation tasks (e.g., ``What is photosynthesis?'' or ``Write a CV''), where sequential scaffolding offers limited diagnostic value. In datasets with sparse skill overlap (e.g., GSM8K, Math-Hard), combinatorial-level analysis is more limited. Importantly, our goal is not to isolate single-skill failures but to reveal how skill \textit{combinations} interact—cumulative scaffolding captures higher-order dependencies and error propagation that isolated evaluations miss, at the cost of less direct attribution to individual sub-skills.

\paragraph{Dataset Filtering Bias.} \framework~retains only examples where the teacher model produces consistent scaffolded variations, which may introduce selection bias. To assess representativeness, we compare model accuracy and semantic cluster distributions between original and retained sets (Appendix~\ref{app:filtering}). For ToT and GSM8K, accuracy differences are small ($<$2.4\%) and cluster proportions are highly correlated ($r > 0.89$), confirming that filtering preserves dataset diversity. For Math-Hard, the larger accuracy gap (9.81\%) and lower correlation ($r = 0.68$) suggest that retained samples may skew easier, and conclusions for this dataset should be interpreted with that caveat.

\paragraph{Generalizability.}
While this work focuses on mathematical reasoning benchmarks with small- and medium-sized models—where compositional skill gaps are most transparent and observable—we see natural directions for future work. Extending \framework~ to broader multi-step reasoning domains, such as tool-calling and complex instruction following, would further demonstrate the generalizability of scaffolded task design as a diagnostic framework.

\section*{Declaration on Generative AI}
During the preparation of this work, the authors used GPT-5 mini to perform grammar and spelling checks, as well as to assist in generating tables and formatting listings. The authors reviewed and edited all content as needed and take full responsibility for the publication's content.

\section*{Acknowledgments}
We would like to thank our colleagues at IBM Research for their support and valuable discussions. In particular, we are grateful to Heiko Ludwig for his guidance and support throughout this project. We also thank the anonymous reviewers for their constructive feedback, which helped improve the quality of this work.

\bibliography{custom}

\appendix
\onecolumn

\section{Background and Related Work}
\label{sec:related_work}

\paragraph{Moving beyond aggregate evaluation.}
Most existing evaluations treat complex tasks as monolithic, and aggregate scores often obscure systematic weaknesses \cite{ribeiro2020beyond, huang2023towards}. Prior work has addressed this by designing more structured evaluations: the MATH dataset introduces fine-grained difficulty categories to better differentiate reasoning ability \cite{hendrycksmath2021}; contrast sets perturb inputs to expose brittle decision boundaries \cite{gardner2020evaluating}; and controlled diagnostic sets reveal that high accuracy can mask reliance on shallow heuristics \cite{mccoy2019right}. Dynabench advocates for multi-dimensional benchmarking beyond single-number evaluations \cite{kiela2021dynabench}, and Robustness Gym unifies robustness testing across diverse transformation conditions \cite{goel2021robustness}. These works collectively motivate structured error analysis that surfaces systematic failure modes hidden by aggregate scores.

\paragraph{Task decomposition for reasoning.}
Another line of work breaks complex tasks into simpler sub-tasks to make intermediate reasoning steps more explicit and configurable \cite{khot2022decomposed, dua2022successive, zhou2022least, prasad2024adapt, jiang2023followbench, laban2025llms}. Chain-of-thought prompting elicits intermediate reasoning traces \cite{wei2022chain}, while least-to-most prompting decomposes problems into sequential subproblems \cite{zhou2022least}. Decomposed prompting modularizes tasks into composable components \cite{khot2022decomposed}, and ADAPT performs as-needed decomposition targeting reasoning bottlenecks \cite{prasad2024adapt}. However, as noted in the introduction, these methods are primarily designed to \textit{improve} performance rather than to systematically diagnose compositional failure modes.

\paragraph{Scaffolding.}
From a cognitive science perspective, competence is not only what a learner can do independently, but also what it can achieve with appropriate guidance \cite{vygotsky1978mind}. Scaffolding refers to this temporary, targeted support that is gradually removed as proficiency develops \cite{van2010scaffolding}. We adapt this principle to LLM evaluation: rather than using scaffolding to improve performance, \framework~uses it as a controlled diagnostic probe to identify which specific skill combinations a model cannot yet perform independently, formalizing this as \textit{scaffolded competence}.

\section{Decomposition Quality}\label{appendix:scaffold_quality}

We conduct a quantitative evaluation of decomposition quality. Table~\ref{tab:scaffold_quality} reports redundancy and coverage across datasets. \textbf{Redundancy} measures whether intermediate steps are repetitive, quantified as the proportion of identical pairs of intermediate answers over all intermediate answers. \textbf{Coverage} measures whether the intermediate steps adequately capture the reasoning process. It is computed via reconstruction success, where the model is given only the intermediate steps and their answers (excluding the original question and final answer) and must reproduce the correct final answer. For generating and evaluating answers, we use Opus 4.6 \cite{anthropic2025opus45} and Sonnet 4.5 \cite{anthropic2025sonnet45}, respectively.

Table~\ref{tab:scaffold_quality} shows low redundancy across all datasets (4.21--8.29\%) and high coverage for ToT and Math-Hard (99.09\% and 99.31\%), with slightly lower coverage on GSM8K (87.52\%), indicating that the generated scaffolds are generally diverse and largely sufficient for reconstructing the original reasoning process.

\begin{table}[H]
\centering
\caption{Quantitative analysis of decomposition quality (Redundancy and Coverage).}
\begin{tabular}{lcc}
\hline
\textbf{Dataset} & \textbf{Redundancy (\%)} & \textbf{Coverage (\%)} \\
\hline
ToT        & 8.29  & 99.09 \\
GSM8K      & 4.21  & 87.52 \\
Math-Hard  & 6.03  & 99.31 \\
\hline
\end{tabular}
\label{tab:scaffold_quality}
\end{table}

\section{Dataset Filtering Validation}\label{app:filtering}

To verify that our filtering process does not introduce bias, we compare model performance and semantic composition between the original and retained datasets using Granite 3.3 8B Instruct.

Table~\ref{tab:filtering_accuracy} shows that accuracy improves slightly after filtering across all three benchmarks, suggesting that removed samples were disproportionately ambiguous or noisy rather than systematically harder or easier.

\begin{table}[H]
\centering
\caption{Performance comparison between original and retained datasets using Granite 3.3 8B Instruct. Accuracy (\%) and raw counts (correct / total).}
\begin{tabular}{lccc}
\hline
\textbf{Dataset} & \textbf{Original} & \textbf{Retained} & \textbf{Diff.} \\
\hline
ToT       & 30.16\% (558/1850)  & 32.45\% (470/1448) & +2.29\% \\
GSM8K     & 88.85\% (1172/1319) & 91.24\% (1073/1176) & +2.39\% \\
Math-Hard & 36.63\% (485/1324)  & 46.44\% (359/773)  & +9.81\% \\
\hline
\end{tabular}

\label{tab:filtering_accuracy}
\end{table}

To further assess whether filtering preserves the semantic distribution of each dataset, we grouped questions into 40 semantic clusters and compared cluster proportions between the original and retained samples. As shown in Table~\ref{tab:filtering_correlation}, all three datasets exhibit high Pearson correlations ($r > 0.67$, $p < 0.001$), confirming that cluster rankings are preserved and that filtering was semantically unbiased.

\begin{table}[H]
\centering
\caption{Correlation of semantic cluster proportions between the original and retained samples. High correlations confirm cluster rankings are preserved and filtering was unbiased.}
\begin{tabular}{lcc}
\hline
\textbf{Dataset} & \textbf{Pearson $r$} & \textbf{$p$-value} \\
\hline
ToT       & 0.9917 & $<$0.001 \\
GSM8K     & 0.8946 & $<$0.001 \\
Math-Hard & 0.6750 & $<$0.001 \\
\hline
\end{tabular}
\label{tab:filtering_correlation}
\end{table}

\section{Cross-Teacher Robustness of Bottleneck Analysis}
\label{app:cross_teacher}

To test whether the identified bottlenecks are artifacts of a particular teacher model's decomposition, we re-instantiate \framework~with a different teacher model (Opus-4.6) and recompute the bottlenecks on the ToT benchmark. Table~\ref{tab:cross_teacher} summarizes the top-5 bottleneck categories identified by each teacher for every student model.

On average, 3.7 out of 5 top bottleneck categories are shared across both teachers, with \texttt{granite-8b} showing perfect agreement (5/5). Even where categories differ, they tend to involve closely related skills (e.g., \textit{Calendar} vs.\ \textit{Calendar/Arithmetic}, \textit{Discrete} vs.\ \textit{Overlap/Discrete}). These results indicate that the identified bottlenecks reflect objective failure points and true skill gaps rather than idiosyncrasies of a single teacher model.

\begin{table*}[t]
\centering
\caption{Top-5 bottleneck analysis using GPT-OSS-120B and Opus-4.6 as teacher models on the ToT benchmark. Numbers in parentheses indicate failure counts. On average, 3.7/5 top bottleneck categories are shared across teachers.}
\small
\begin{tabular}{lp{4.5cm}p{4.5cm}p{3cm}c}
\hline
\textbf{Model} & \textbf{GPT-OSS-120B (paper)} & \textbf{Opus-4.6 (new)} & \textbf{Shared} & \textbf{Overlap} \\
\hline
qwen-3b &
  Overlap/Discrete (56), Unit (45), Natural (43), Natural/Calendar (39), Calendar/Arithmetic (31) &
  Natural (78), Overlap/Discrete (49), Natural/Calendar (36), Calendar (36), Discrete (36) &
  Overlap/Discrete, Natural, Natural/Calendar & 3/5 \\
\hline
qwen-7b &
  Natural (52), Overlap/Discrete (49), Unit (48), Calendar (40), Overlap (39) &
  Discrete (55), Natural (52), Overlap/Discrete (48), Calendar (45), Unit (34) &
  Natural, Overlap/Discrete, Unit, Calendar & 4/5 \\
\hline
llama-3b &
  Overlap/Discrete (58), Natural (42), Unit/Arithmetic (36), Natural/Calendar (34), Calendar (28) &
  Natural (56), Overlap/Discrete (41), Natural/Calendar (37), Discrete (33), Normalization (21) &
  Overlap/Discrete, Natural, Natural/Calendar & 3/5 \\
\hline
llama-8b &
  Overlap/Discrete (59), Unit/Arithmetic (57), Natural (44), Natural/Calendar (41), Calendar (38) &
  Natural (54), Discrete (52), Overlap/Discrete (49), Calendar (40), Normalization (34) &
  Overlap/Discrete, Natural, Calendar & 3/5 \\
\hline
granite-2b &
  Overlap/Discrete (66), Unit/Arithmetic (54), Natural/Calendar (43), Calendar (34), Calendar/Arithmetic (34) &
  Natural (53), Natural/Calendar (37), Overlap/Discrete (35), Calendar (33), Unit/Arithmetic (29) &
  Overlap/Discrete, Unit/Arithmetic, Natural/Calendar, Calendar & 4/5 \\
\hline
granite-8b &
  Natural (59), Overlap/Discrete (59), Unit/Arithmetic (53), Natural/Calendar (50), Unit (49) &
  Natural (74), Unit (54), Unit/Arithmetic (52), Natural/Calendar (51), Overlap/Discrete (50) &
  Natural, Overlap/Discrete, Unit/Arithmetic, Natural/Calendar, Unit & 5/5 \\
\hline
\end{tabular}
\label{tab:cross_teacher}
\end{table*}

\section{Ablation: Scaffolding Effect vs.\ Information Leakage}
\label{app:ablation_leakage}

To verify that scaffolding gains stem from supporting reasoning steps rather than simply reducing task complexity, we conduct a control experiment on ToT Arithmetic with three models. We compare three conditions:

\begin{itemize}
    \item \textbf{Question (baseline):} Original task, no scaffolding. By construction, these are questions the model answered incorrectly (0\% accuracy).
    \item \textbf{Scaffolding:} The minimum scaffolded level at which the model succeeds is selected, with correct intermediate answers injected (100\% accuracy by construction).
    \item \textbf{Control A (placeholders only):} Injected values are replaced with non-informative placeholders (e.g., \texttt{[VALUE]}), preserving task structure without revealing intermediate answers.
\end{itemize}

Table~\ref{tab:ablation_leakage} shows the results. Under Control A, accuracy drops to 7.6--14.4\% (avg.\ 11.8\%) across all three models, closely matching the zero baseline. This confirms that performance gains under scaffolding are driven by the injected intermediate values—not by task restructuring or linguistic simplification alone.

\begin{table}[H]
\centering
\caption{Ablation results on ToT Arithmetic (n=538, 515, 616 respectively). Replacing injected values with placeholders drops accuracy to near-baseline (avg.\ 11.8\%), confirming that scaffolding gains are driven by intermediate answer injection, not task restructuring.}
\begin{tabular}{lccc}
\hline
\textbf{Condition} & \textbf{Qwen2.5-7B} & \textbf{Llama-3.1-8B} & \textbf{Granite-3.3-8B} \\
\hline
Question (baseline)      & 0\%   & 0\%    & 0\%    \\
Scaffolding              & 100\% & 100\%  & 100\%  \\
Control A (placeholders) & 7.6\% & 14.4\% & 13.3\% \\
\hline
\end{tabular}
\label{tab:ablation_leakage}
\end{table}

\section{Number of Clusters and Skills}\label{appendix:cluster_skill}

\begin{figure}[H]
    \centering
    \begin{subfigure}{\textwidth}
        \centering
        \includegraphics[width=\textwidth]{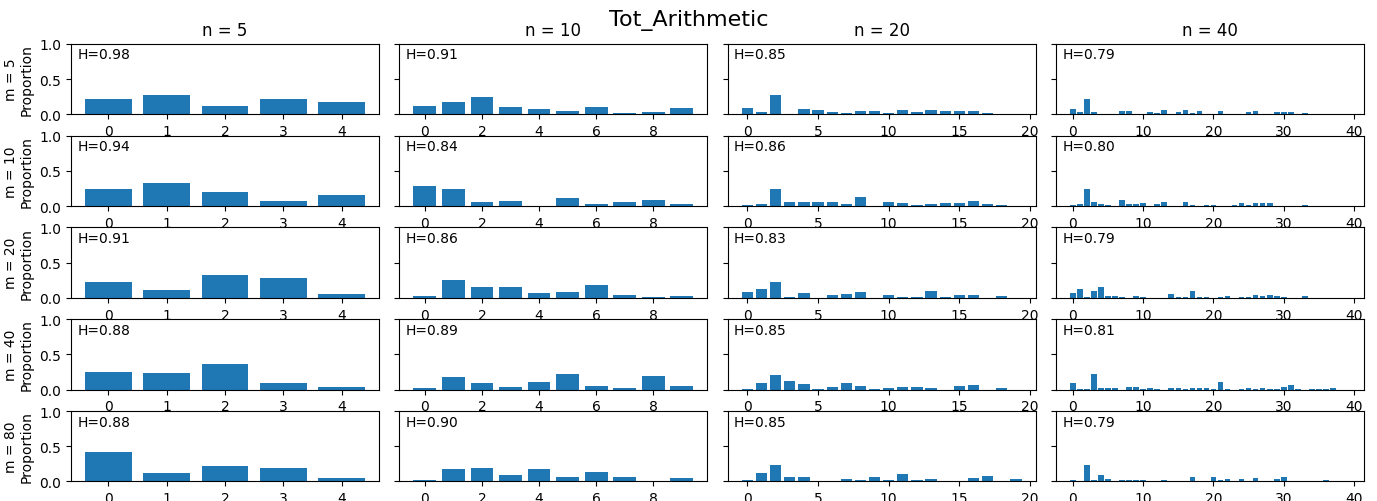}
        \caption{ToT Arithmetic}
    \end{subfigure}

    \vspace{0.6em}

    \begin{subfigure}{\textwidth}
        \centering
        \includegraphics[width=\textwidth]{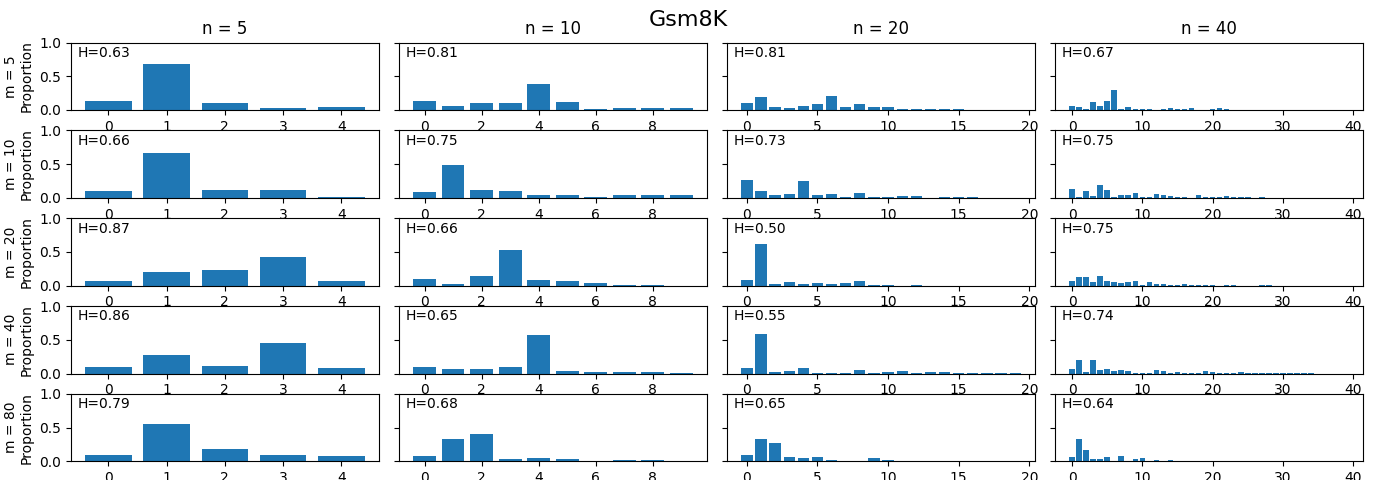}
        \caption{GSM8K}
    \end{subfigure}

    \vspace{0.6em}

    \begin{subfigure}{\textwidth}
        \centering
        \includegraphics[width=\textwidth]{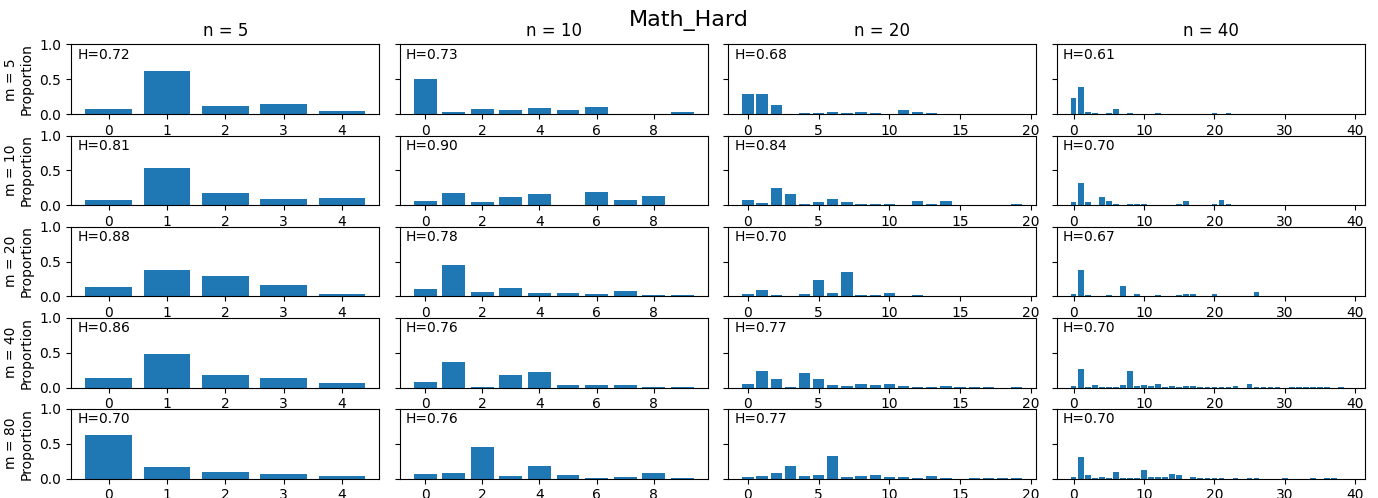}
        \caption{Math-Hard}
    \end{subfigure}

    \vspace{0.6em}

    \caption{Skill Distribution Across (m, n) Settings. X-axis skill ID each distinct skill and y axis represent the ratio of the usage. H represents entropy.}
    \label{fig:skill_distribution}
\end{figure}

\begin{figure}[H]
    \centering
    \caption{Other-category coverage across $(M, N)$ configurations for each dataset. Each line represents a different number of clusters ($M$); the y-axis shows the proportion of tasks assigned to the catch-all \emph{other} skill. Lower values indicate better task coverage by the defined skill set.}

    \begin{subfigure}{0.32\textwidth}
        \centering
        \includegraphics[width=\textwidth]{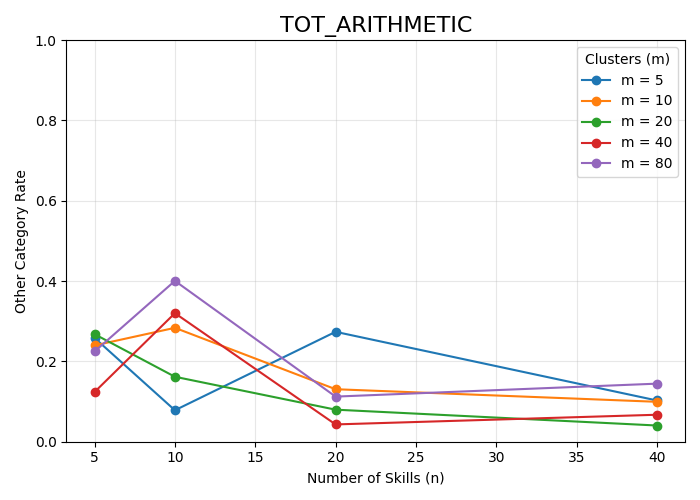}
        \caption{ToT Arithmetic}
    \end{subfigure}
    \hfill
    \begin{subfigure}{0.32\textwidth}
        \centering
        \includegraphics[width=\textwidth]{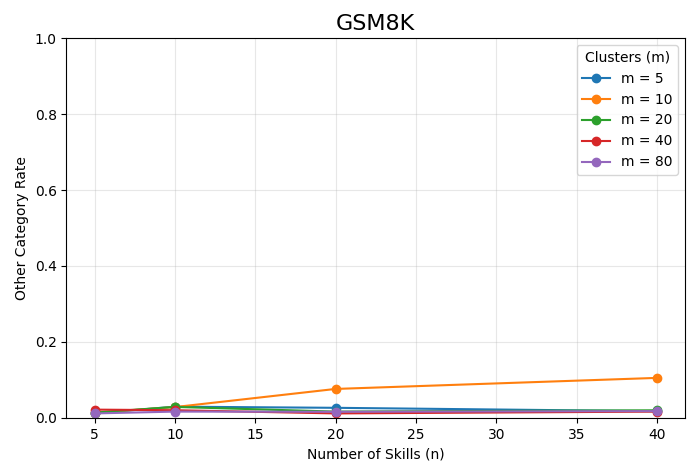}
        \caption{GSM8K}
    \end{subfigure}
    \hfill
    \begin{subfigure}{0.32\textwidth}
        \centering
        \includegraphics[width=\textwidth]{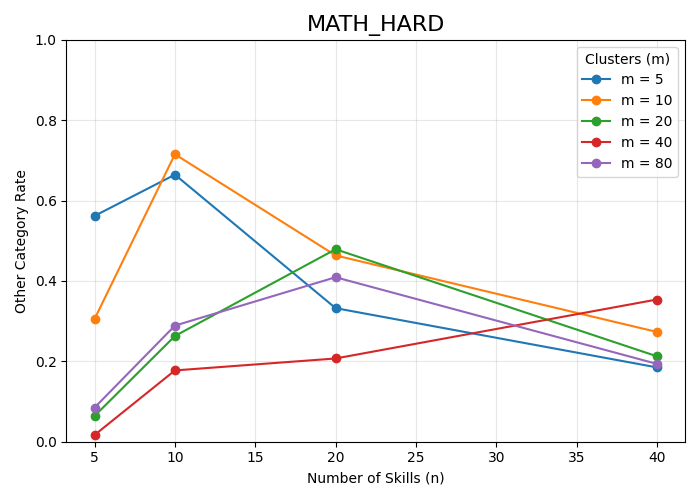}
        \caption{Math-Hard}
    \end{subfigure}

    \label{fig:other_cateogry_rate}
\end{figure}

We select the number of representative task clusters ($M$) and higher-level skills ($N$) via systematic search over $M \in \{5, 10, 20, 40, 80\}$ and $N \in \{5, 10, 20, 40\}$, yielding 20 combinations per dataset. We evaluate each configuration on four criteria:

\begin{itemize}[noitemsep]
    \item \textbf{Skill distribution} (Figure~\ref{fig:skill_distribution}): entropy of skill usage. Higher entropy indicates more balanced coverage.
    \item \textbf{Other-category coverage} (Figure~\ref{fig:other_cateogry_rate}): proportion of tasks assigned to the catch-all \emph{other} skill. Lower is better.
    \item \textbf{Skill granularity} (Figures~\ref{fig:skill_granularity_tot},~\ref{fig:skill_granularity_gsm8k},~\ref{fig:skill_granularity_math_hard}): whether skills are too broad or overly task-specific.
    \item \textbf{Skill overlap} (Table~\ref{tab:skill_overlap_all}): cosine similarity between skills. High overlap indicates redundancy.
    
\end{itemize}

The final skill lists are provided in Appendix~\ref{appendix:skill_list}.

\subsection{ToT Arithmetic}\label{appendix:cluster_skill_tot}
\begin{itemize}
    \item \textbf{Coverage.} Increasing $N$ consistently reduces the \emph{other} rate, but gains diminish beyond $N=20$. The lowest rates occur at $M \in \{20, 40\}$.
    \item \textbf{Granularity.} At $N=5$, a single skill (e.g., ``duration arithmetic'') subsumes multiple distinct sub-tasks. At $N=10$, skills improve but remain coarse. At $N=20$, granularity is appropriate: skills such as ``overlap and intersection of time intervals'' and ``discrete slot counting under constraints'' map cleanly to individual reasoning steps. $N=40$ offers similar quality without clear benefit.
    \item \textbf{Overlap.} Most skills remain distinct (low mean similarity). Notable redundancy appears at $(M,N) \in \{(10,40),(20,40),(40,40)\}$, where a non-trivial fraction of pairs exceed similarity $\geq 0.78$, favoring smaller $N$.
    \item  \textbf{Distribution.} Entropy decreases as $N$ grows, producing sparse long-tailed distributions at large $N$. Varying $M$ has minimal effect.
    \item \textbf{Selection: $M=40$, $N=20$} — achieves low uncovered-task rates, interpretable skill granularity, low redundancy, and balanced skill usage.
\end{itemize}

\subsection{GSM8K}
\begin{itemize}
    \item \textbf{Coverage.} The \emph{other} rate is consistently low across most configurations. Smaller $M$ ($\leq 10$) degrades at higher $N$; $M \in \{20,40\}$ is stable. Increasing $M$ to 80 yields no gain.

    \item \textbf{Granularity.} Skills at $N \leq 20$ remain too coarse (e.g., ``sequential quantity tracking'' covers multiple sub-tasks). At $N=40$, skills reach appropriate specificity (e.g., ``determining remaining quantity after consumption and distribution'', ``using unit rates to compute totals'').

    \item \textbf{Overlap.} Only $(M,N)=(10,40)$ shows notable redundancy ($\geq 0.78$); all other configurations are distinctive.

    \item \textbf{Selection: $M=40$, $N=40$} — required for sufficient granularity on the diverse arithmetic reasoning patterns in GSM8K.
\end{itemize}

\subsection{Math-Hard}
\begin{itemize}
    \item \textbf{Coverage.} The \emph{other} rate is more variable across configurations. The lowest rates occur at $(M,N) \in \{(20,5),(40,5),(80,5),(40,10),(40,20),(5,40),(80,40)\}$.

    \item \textbf{Granularity.} $N \geq 20$ provides sufficient specificity (Figure~\ref{fig:skill_granularity_math_hard}).

    \item \textbf{Overlap.} All tested configurations show low redundancy (Table~\ref{tab:skill_overlap_all}).

    \item \textbf{Selection: $M=40$, $N=20$} — balances coverage, granularity, and distinctiveness for competition-level math problems.
\end{itemize}

\newpage

\begin{figure}[H]
  \centering
  \begin{tcolorbox}[
      enhanced,
      fontupper=\small,
      colback=lightgray,
      colframe=darkgray,
      boxrule=0.8pt,
      arc=3pt,
      title={\small\textbf{Skill Granularity Comparison (ToT)}},
      coltitle=white,
      colbacktitle=darkgray,
      boxsep=2pt,
      left=4pt,
      right=4pt,
      top=2pt,
      bottom=2pt,
      before upper={\setlength{\parskip}{1pt}},
  ]
  \colorbox{lightblue}{\parbox{\dimexpr\linewidth-2\fboxsep}{%
  {\color{promptblue}\textbf{Original Problem}}

  The availability of Brandon is 11 to 12:30 and also from 2:30 to 4:30 and the availability of Levi is from 9 to 11:30 and also from 4:30 to 5. How many possibilities are there for the meeting time if the meeting has to start on the hour or half hour and must be 30 minutes long. Format your answer as a JSON. Eg. JSON = {\"explanation\": <your step by step solution>, \"answer\":<number>}. Do not include units in the JSON value.
  }}

  \vspace{2pt}

  \colorbox{white}{\parbox{\dimexpr\linewidth-2\fboxsep}{%
  {\color{black}\textbf{Sub-Tasks}} 

   \{"sub-task": "Find the overlapping time slots between Brandon and Levi"\}
   
   \{"sub-task": "Determine the possible 30-minute meeting slots within the overlapping time slots that start on the hour or half hour"\}
   
   \{"sub-task": "Count the total number of possible 30-minute meeting slots"\}
   
   \{"sub-task": "Format the answer as a JSON"\}

  }}

  \vspace{2pt}

  \colorbox{white}{\parbox{\dimexpr\linewidth-2\fboxsep}{%
  {\color{black}\textbf{Mapped Skills (m=40, n=5)}}
  
    \{"skill": "Duration arithmetic"\}
    
    \{"skill": "Duration arithmetic"\}
    
    \{"skill": "Duration arithmetic"\}
    
    \{"skill": "Structured JSON reporting"\}

  }
  }

  \vspace{2pt}

  \colorbox{white}{\parbox{\dimexpr\linewidth-2\fboxsep}{%
  {\color{black}\textbf{Mapped Skills (m=40, n=10)}}
  
    \{"skill": "Constraint-driven scheduling slot enumeration"\}
    
    \{"skill": "Constraint-driven scheduling slot enumeration"\}
    
    \{"skill": "Constraint-driven scheduling slot enumeration"\}
    
    \{"skill": "Constraint-driven scheduling slot enumeration"\}

  }
  }

  \vspace{2pt}

  \colorbox{white}{\parbox{\dimexpr\linewidth-2\fboxsep}{%
  {\color{black}\textbf{Mapped Skills (m=40, n=20)}}
  
    \{"skill": "Overlap and intersection of time intervals"\}
    
    \{"skill": "Discrete slot counting under constraints"\}
    
    \{"skill": "Discrete slot counting under constraints"\}
    
    \{"skill": "Structured JSON output generation"\}

  }
  }

    \vspace{2pt}

  \colorbox{white}{\parbox{\dimexpr\linewidth-2\fboxsep}{%
  {\color{black}\textbf{Mapped Skills (m=40, n=40)}}
  
    \{"skill": "Finding overlapping free intervals among multiple schedules"\}
    
    \{"skill": "Counting feasible meeting slots with granularity constraints"\}
    
    \{"skill": "Counting feasible meeting slots with granularity constraints"\}
    
    \{"skill": "Formatting results into a prescribed JSON structure"\}

  }
  }

  \end{tcolorbox}
  \caption{ToT Skill Granularity m=40 and where n=5, 10, 20, 40.}
  \label{fig:skill_granularity_tot}
\end{figure}

\begin{figure}[H]
  \centering
  \begin{tcolorbox}[
      enhanced,
      fontupper=\small,
      colback=lightgray,
      colframe=darkgray,
      boxrule=0.8pt,
      arc=3pt,
      title={\small\textbf{Skill Granularity Comparison (GSM8K)}},
      coltitle=white,
      colbacktitle=darkgray,
      boxsep=2pt,
      left=4pt,
      right=4pt,
      top=2pt,
      bottom=2pt,
      before upper={\setlength{\parskip}{1pt}},
  ]
  \colorbox{lightblue}{\parbox{\dimexpr\linewidth-2\fboxsep}{%
  {\color{promptblue}\textbf{Original Problem}}

  Janet's ducks lay 16 eggs per day. She eats three for breakfast every morning and bakes muffins for her friends every day with four. She sells the remainder at the farmers' market daily for \$2 per fresh duck egg. How much in dollars does she make every day at the farmers' market?
  }}

  \vspace{2pt}

  \colorbox{white}{\parbox{\dimexpr\linewidth-2\fboxsep}{%
  {\color{black}\textbf{Sub-Tasks}} 

   \{"sub-task": "Determine the total number of eggs produced each day"\}

   \{"sub-task": "Subtract the three eggs Janet eats for breakfast from the total"\}

   \{"sub-task": "Subtract the four eggs used for muffins from the remaining eggs"\}

   \{"sub-task": "Calculate the number of eggs left after breakfast and muffins, which are sold"\}

   \{"sub-task": "Multiply the number of eggs sold by \$2 to find the daily earnings at the farmers' market"\}

  }}

  \vspace{2pt}

  \colorbox{white}{\parbox{\dimexpr\linewidth-2\fboxsep}{%
  {\color{black}\textbf{Mapped Skills (m=40, n=5)}}
  
    \{"skill": "Quantitative language comprehension"\}

    \{"skill": "Sequential quantity tracking"\}

    \{"skill": "Sequential quantity tracking"\}

    \{"skill": "Sequential quantity tracking"\}

    \{"skill": "Unit conversion and rate reasoning"\}

  }
  }

  \vspace{2pt}

  \colorbox{white}{\parbox{\dimexpr\linewidth-2\fboxsep}{%
  {\color{black}\textbf{Mapped Skills (m=40, n=10)}}
  
    \{"skill": "Narrative-to-Math Translation"\}

    \{"skill": "Sequential Quantity Tracking"\}

    \{"skill": "Sequential Quantity Tracking"\}

    \{"skill": "Sequential Quantity Tracking"\}

    \{"skill": "Rate and Proportional Reasoning"\}

  }
  }

  \vspace{2pt}

  \colorbox{white}{\parbox{\dimexpr\linewidth-2\fboxsep}{%
  {\color{black}\textbf{Mapped Skills (m=40, n=20)}}
  
    \{"skill": "Word-problem interpretation"\}

    \{"skill": "Basic arithmetic operations"\}

    \{"skill": "Basic arithmetic operations"\}

    \{"skill": "Sequential quantity tracking"\}

    \{"skill": "Monetary cost and change calculation"\}
    
  }
  }

    \vspace{2pt}

  \colorbox{white}{\parbox{\dimexpr\linewidth-2\fboxsep}{%
  {\color{black}\textbf{Mapped Skills (m=40, n=40)}}
  
    \{"skill": "Extracting quantitative data from narrative text"\}

    \{"skill": "Sequential quantity tracking"\}

    \{"skill": "Sequential quantity tracking"\}

    \{"skill": "Determining remaining quantity after consumption and distribution"\}

    \{"skill": "Using unit rates to compute totals"\}

  }
  }

  \end{tcolorbox}
  \caption{GSM8K Skill Granularity m=40 and where n=5, 10, 20, 40.}
  \label{fig:skill_granularity_gsm8k}
\end{figure}

\begin{figure}[H]
  \centering
  \begin{tcolorbox}[
      enhanced,
      fontupper=\small,
      colback=lightgray,
      colframe=darkgray,
      boxrule=0.8pt,
      arc=3pt,
      title={\small\textbf{Skill Granularity Comparison (Math-Hard)}},
      coltitle=white,
      colbacktitle=darkgray,
      boxsep=2pt,
      left=4pt,
      right=4pt,
      top=2pt,
      bottom=2pt,
      before upper={\setlength{\parskip}{1pt}},
  ]
  \colorbox{lightblue}{\parbox{\dimexpr\linewidth-2\fboxsep}{%
  {\color{promptblue}\textbf{Original Problem}}

  What is the smallest real number $x$ in the domain of the function $g(x) = \\sqrt{(x-3)^2-(x-8)^2}~?$
  }}

  \vspace{2pt}

  \colorbox{white}{\parbox{\dimexpr\linewidth-2\fboxsep}{%
  {\color{black}\textbf{Sub-Tasks}} 

   \{"sub-task": "Set the radicand $(x-3)^2 - (x-8)^2$ greater than or equal to zero"\}

   \{"sub-task": "Apply the difference of squares formula to rewrite the radicand as $[(x-3)-(x-8)]·[(x-3)+(x-8)]$"\}

   \{"sub-task": "Simplify each factor to obtain a linear expression multiplied by a constant"\}

   \{"sub-task": "Solve the resulting linear inequality for $x$ and identify the smallest real $x$ that satisfies it"\}
   }}
  \vspace{2pt}

  \colorbox{white}{\parbox{\dimexpr\linewidth-2\fboxsep}{%
  {\color{black}\textbf{Mapped Skills (m=80, n=5)}}
  
    \{"skill": "Algebraic manipulation of equations and unknown coefficients"\}

    \{"skill": "Algebraic manipulation of equations and unknown coefficients"\}

    \{"skill": "Algebraic manipulation of equations and unknown coefficients"\}

    \{"skill": "Algebraic manipulation of equations and unknown coefficients"\}
  }
  }

  \vspace{2pt}

  \colorbox{white}{\parbox{\dimexpr\linewidth-2\fboxsep}{%
  {\color{black}\textbf{Mapped Skills (m=80, n=10)}}
  
    \{"skill": "Translating problem conditions into algebraic equations"\}

    \{"skill": "Factoring polynomials and using the factor theorem"\}

    \{"skill": "Factoring polynomials and using the factor theorem"\}

    \{"skill": "Solving systems of linear equations for unknown parameters"\}

  }
  }

  \vspace{2pt}

  \colorbox{white}{\parbox{\dimexpr\linewidth-2\fboxsep}{%
  {\color{black}\textbf{Mapped Skills (m=80, n=20)}}
  
    \{"skill": "Domain analysis for radicals and rational functions"\}

    \{"skill": "Factoring and completing the square"\}

    \{"skill": "Factoring and completing the square"\}

    \{"skill": "Domain analysis for radicals and rational functions"\}

  }
  }

    \vspace{2pt}

  \colorbox{white}{\parbox{\dimexpr\linewidth-2\fboxsep}{%
  {\color{black}\textbf{Mapped Skills (m=80, n=40)}}
  
    \{"skill": "Determining domains of radical and rational functions"\}

    \{"skill": "Factoring differences of squares and other algebraic identities"\}

    \{"skill": "Factoring differences of squares and other algebraic identities"\}

    \{"skill": "Determining domains of radical and rational functions"\}

  }
  }

  \end{tcolorbox}
  \caption{Math-Hard Skill Granularity m=80 and where n=5, 10, 20, 40.}
  \label{fig:skill_granularity_math_hard}
\end{figure}

\newpage

\begin{table*}[t]
\centering
\small
\caption{Intra-skill semantic overlap statistics across different skill granularities.}
\label{tab:skill_overlap_all}
\begin{tabular}{cc|rr|rr|rr}
\hline
$m$ & $n$ & \multicolumn{2}{c|}{ToT Arithmetic} & \multicolumn{2}{c|}{GSM8K} & \multicolumn{2}{c}{Math-Hard} \\
    &     & Mean Sim. & \% $\geq$ 0.78 & Mean Sim. & \% $\geq$ 0.78 & Mean Sim. & \% $\geq$ 0.78 \\
\hline
5  & 5  & 0.370 & 0.0  & 0.322 & 0.0  & 0.131 & 0.0  \\
5  & 10 & 0.345 & 0.0  & 0.268 & 0.0  & 0.174 & 0.0  \\
5  & 20 & 0.306 & 0.0  & 0.302 & 0.0  & 0.151 & 0.0  \\
5  & 40 & 0.276 & 0.0  & 0.282 & 0.0  & 0.153 & 0.0  \\
10 & 5  & 0.409 & 0.0  & 0.346 & 0.0  & 0.203 & 0.0  \\
10 & 10 & 0.326 & 0.0  & 0.305 & 0.0  & 0.194 & 0.0  \\
10 & 20 & 0.278 & 0.0  & 0.260 & 0.0  & 0.205 & 0.0  \\
10 & 40 & 0.289 & 0.0  & 0.280 & 13.5 & 0.192 & 0.0  \\
20 & 5  & 0.349 & 0.0  & 0.426 & 0.0  & 0.272 & 0.0  \\
20 & 10 & 0.328 & 0.0  & 0.303 & 0.0  & 0.219 & 0.0  \\
20 & 20 & 0.283 & 0.0  & 0.265 & 0.0  & 0.198 & 0.0  \\
20 & 40 & 0.302 & 38.5 & 0.300 & 0.0  & 0.172 & 0.0  \\
40 & 5  & 0.257 & 0.0  & 0.302 & 0.0  & 0.332 & 0.0  \\
40 & 10 & 0.337 & 0.0  & 0.262 & 0.0  & 0.219 & 0.0  \\
40 & 20 & 0.301 & 0.0  & 0.308 & 0.0  & 0.170 & 0.0  \\
40 & 40 & 0.301 & 12.8 & 0.293 & 0.0  & 0.169 & 0.0  \\
80 & 5  & 0.311 & 0.0  & 0.351 & 0.0  & 0.227 & 0.0  \\
80 & 10 & 0.358 & 0.0  & 0.328 & 0.0  & 0.243 & 0.0  \\
80 & 20 & 0.291 & 0.0  & 0.261 & 0.0  & 0.173 & 0.0  \\
80 & 40 & 0.328 & 0.0  & 0.235 & 0.0  & 0.151 & 0.0  \\
\hline
\end{tabular}
\end{table*}

\newpage

\section{Skill List}\label{appendix:skill_list}

\subsection{20 Skills in ToT Arithmetic}
\begin{enumerate}[noitemsep, topsep=2pt]
    \item \textbf{Natural-language date/time parsing}: Extract dates, times, and durations from varied textual expressions and formats.
    \item \textbf{Unit conversion for time spans}: Translate hours, minutes, seconds, days, weeks, months, and years into a common unit (e.g., total seconds) and back.
    \item \textbf{Arithmetic on durations}: Add, subtract, multiply, or divide time intervals, including scaling rates to larger quantities.
    \item \textbf{Calendar arithmetic}: Compute new dates by adding or removing days, weeks, months, or years, respecting month lengths and leap years.
    \item \textbf{Leap-year and month-length handling}: Determine the correct number of days in February and other months when performing date calculations.
    \item \textbf{Time-zone conversion}: Apply UTC offsets to convert times between zones and adjust the resulting day if needed.
    \item \textbf{Day-of-week determination}: Find the weekday for a given date or after a specified offset using modular arithmetic on a 7-day cycle.
    \item \textbf{Relative-date reasoning}: Interpret statements like ``X days before/after Y'' or ``X weeks earlier'' to locate the target date.
    \item \textbf{Normalization of incomplete timestamps}: Fill missing components (e.g., assume 00:00:00 for absent time) to create comparable datetime values.
    \item \textbf{Chronological ordering of events}: Sort a set of dates/times ascending or descending to identify earliest or latest occurrences.
    \item \textbf{Extremum identification}: Locate the maximum or minimum date/time among a collection (e.g., latest exam, earliest activity).
    \item \textbf{Overlap and intersection of time intervals}: Determine common free periods among multiple schedules and compute their length.
    \item \textbf{Discrete slot counting under constraints}: Enumerate possible meeting or event slots that satisfy granularity rules (e.g., start on the hour or half-hour).
    \item \textbf{Midnight and date-boundary wrap-around}: Correctly handle intervals that cross midnight or span month/year boundaries.
    \item \textbf{Multi-format date conversion}: Translate between representations such as mm/dd/yyyy, yyyy-mm-dd, dd-Mon-yyyy, etc.
    \item \textbf{Era linearization}: Convert BC and AD years to a single numeric timeline for comparison.
    \item \textbf{Parsing and applying weekly cycles}: Use recurring periods (e.g., every 19 days) to compute next occurrence from a reference date.
    \item \textbf{Rate-based scaling}: Derive per-unit time from a given total and apply it to a different quantity (e.g., time per box).
    \item \textbf{Structured JSON output generation}: Assemble explanations and computed values into the required JSON schema with correct field names.
    \item \textbf{Ambiguity resolution in temporal language}: Disambiguate phrases like ``before noon'', ``same day'', or ``earliest possible time'' to select the appropriate interpretation.
\end{enumerate}

\subsection{40 Skills in GSM8K}
\begin{enumerate}[noitemsep, topsep=2pt]
    \item \textbf{Extracting quantitative data from narrative text}: Identify all numbers, entities, and relationships described in a word problem.
    \item \textbf{Converting percentages to fractions or decimals}: Translate percentage statements into usable numeric forms for calculation.
    \item \textbf{Performing operations with fractions}: Add, subtract, multiply, and divide fractional quantities accurately.
    \item \textbf{Calculating a percentage of a quantity}: Apply a percent to find part-of-a-whole values.
    \item \textbf{Proportional reasoning}: Set up multiplicative relationships based on comparative language (e.g., ``twice as many'', ``half as much'').
    \item \textbf{Formulating and solving simple linear equations}: Turn a verbal condition into an equation and isolate the variable.
    \item \textbf{Formulating and solving basic inequalities}: Use ``at least'', ``no more than'', etc., to create and solve inequality constraints.
    \item \textbf{Using unit rates to compute totals}: Multiply a per-unit value (price, speed, etc.) by a quantity to obtain a total amount.
    \item \textbf{Multi-digit multiplication and division}: Carry out accurate arithmetic with two- or three-digit numbers.
    \item \textbf{Sequential quantity tracking}: Update a running total through successive additions and subtractions.
    \item \textbf{Converting between time units}: Change minutes to hours or vice versa for combined-time calculations.
    \item \textbf{Area calculation for rectangles and aggregation}: Compute length times width and sum multiple areas to find a total.
    \item \textbf{Total cost from per-item price and quantity}: Multiply unit price by number of items and sum across categories.
    \item \textbf{Finding change or remaining amount after purchases}: Subtract total expense from given money to obtain leftover funds.
    \item \textbf{Rounding up when dividing into packs}: Determine the smallest whole number of packs needed to cover a quantity.
    \item \textbf{Applying average rates to determine total output or time}: Use a rate over a period to compute overall amount.
    \item \textbf{Profit calculation}: Subtract total expenses from total sales to obtain net gain.
    \item \textbf{Applying discount percentages to original prices}: Reduce a price by a given percent and compute the discounted amount.
    \item \textbf{Computing savings from price differences over time}: Multiply per-unit savings by quantity and by number of periods.
    \item \textbf{Inventory tracking over multiple days}: Account for daily usage, additions, and removals to find initial or final stock.
    \item \textbf{Interpreting ``more than'' or ``less than'' relationships}: Translate comparative statements into addition or subtraction equations.
    \item \textbf{Translating ``per'' statements into multiplication}: Convert expressions like ``\$X per hour'' into a product of rate and time.
    \item \textbf{Mixed-operation word problems with order of operations}: Apply PEMDAS when several operations appear together.
    \item \textbf{Substitution to solve for unknowns}: Replace a variable with its known value to evaluate an expression.
    \item \textbf{Consistent handling of mixed measurement units}: Keep units (gallons, inches, miles, etc.) uniform throughout calculations.
    \item \textbf{Calculating totals from grouped items}: Multiply group size by members per group and sum across groups.
    \item \textbf{Determining remaining quantity after consumption and distribution}: Subtract used and given-away amounts from an initial total.
    \item \textbf{Equal sharing of a total among participants}: Divide a quantity evenly to find each person's share.
    \item \textbf{Using ``at least'' conditions to find minimum required values}: Set up an inequality and solve for the smallest feasible number.
    \item \textbf{Converting dozens to individual units}: Multiply a dozen count by 12 to obtain the exact number of items.
    \item \textbf{Summing contributions from multiple sources}: Add separate amounts (e.g., earnings, donations) to obtain a combined figure.
    \item \textbf{Multiplicative scaling for unknown quantities}: Represent ``n times'' relationships as multiplication in equations.
    \item \textbf{Distance calculation from speed and time segments}: Multiply speed by each time interval and sum the distances.
    \item \textbf{Total weight determination}: Sum the weights of all objects carried or listed.
    \item \textbf{Using difference statements to isolate unknown quantities}: Rearrange ``total minus known equals unknown'' relationships to solve.
    \item \textbf{Interpreting ``total of'' statements to set up equations}: Translate ``the total is X'' into an equation linking component parts.
    \item \textbf{Applying ``per pack'' pricing to compute overall cost}: Multiply number of packs by price per pack, accounting for partial packs if needed.
    \item \textbf{Calculating weekly or monthly earnings from hourly wages}: Multiply hourly rate by total hours worked in the period.
    \item \textbf{Solving mixture problems with weighted averages}: Use weighted-average formulas to find unknown component values.
    \item \textbf{Budget allocation across multiple items under constraints}: Distribute a fixed amount among purchases while respecting given limits.
\end{enumerate}

\subsection{20 Skills in Math-Hard}
\begin{enumerate}[noitemsep, topsep=2pt]
    \item \textbf{Translating word problems into algebraic statements}: Extract quantities, relationships, and constraints from prose and express them as equations or inequalities.
    \item \textbf{Multi-step arithmetic with integers and fractions}: Perform sequences of additions, subtractions, multiplications, and divisions accurately, including reduction of fractions.
    \item \textbf{Combinatorial counting}: Use binomial coefficients and factorial reasoning to enumerate selections, arrangements, and distributions.
    \item \textbf{Inclusion-exclusion reasoning}: Account for overlapping cases by adding and subtracting intersecting counts to obtain correct totals.
    \item \textbf{Probability via counting}: Compute probabilities by determining the number of favorable outcomes divided by total equally likely outcomes.
    \item \textbf{Solving linear equations and systems}: Isolate variables, substitute, and use elimination or matrix methods to find unknown values.
    \item \textbf{Quadratic equation techniques}: Factor, complete the square, or apply the quadratic formula to find real or complex roots.
    \item \textbf{Converting repeating decimals to fractions}: Set up algebraic equations for repeating blocks, solve for the unknown, and simplify to lowest terms.
    \item \textbf{Arithmetic and geometric series analysis}: Identify first term and common difference/ratio, use sum formulas, and test convergence for infinite series.
    \item \textbf{Modular arithmetic and congruence solving}: Work with residues, solve linear congruences, and apply the Chinese Remainder Theorem when needed.
    \item \textbf{GCD and LCM via prime factorization}: Decompose integers into primes to compute greatest common divisor and least common multiple efficiently.
    \item \textbf{Absolute-value inequality manipulation}: Split into casewise linear inequalities, solve each case, and intersect solution sets.
    \item \textbf{Domain and range of functions}: Impose non-negativity, non-zero denominator, and piecewise analysis to describe permissible inputs and outputs.
    \item \textbf{Completing the square for optimization}: Rewrite quadratic expressions as a perfect square plus constant to locate minima or maxima.
    \item \textbf{Calculus-based optimization}: Differentiate area, volume, or other expressions, set derivatives to zero, and verify extremal values.
    \item \textbf{Vector orthogonality via dot and cross products}: Compute cross products, take dot products, and set results to zero to enforce perpendicularity.
    \item \textbf{Rayleigh quotient and eigenvalue maximization}: Recognize quadratic forms, relate them to eigenvalues, and use the largest eigenvalue to obtain maximal ratios.
    \item \textbf{Solving higher-degree polynomials}: Apply substitutions, depressed-cubic forms, and Cardano's formula to obtain exact roots.
    \item \textbf{Lattice-point counting under distance constraints}: Use integer solutions of circle equations ($x^2 + y^2 = r^2$) to enumerate points satisfying a given distance.
    \item \textbf{Symmetry counting with Burnside's Lemma}: Identify group actions (rotations, reflections), count fixed configurations under each, and average to obtain distinct arrangements.
\end{enumerate}

\newpage
\section{Prompts}

\begin{figure}[H]
  \centering
  \begin{tcolorbox}[
    enhanced,
    fontupper=\small,
    colback=lightgray,
    colframe=darkgray,
    boxrule=0.8pt,
    arc=3pt,
    title={\small\textbf{Sub-Task Decomposition}},
    coltitle=white,
    colbacktitle=darkgray,
    boxsep=2pt,
    left=4pt,
    right=4pt,
    top=2pt,
    bottom=2pt,
    before upper={\setlength{\parskip}{1pt}},
    listing only,           
    listing options={       
        basicstyle=\ttfamily\small,
        breaklines=true
    }
]

  \colorbox{lightgray}{\parbox{\dimexpr\linewidth-2\fboxsep}{%

You will be given a question that requires multiple reasoning or computational steps.
\\
Your task is to break down the instruction into explicit, step-by-step sequential segments that detail the actions needed to answer the question. Each segment should represent a distinct actionable operation.
\\ \\
Rules: \\
- Each segment must represent a concrete reasoning or computational action.\\
- Each segment must yield a concrete intermediate result.\\
- Each segment must be non-overlapping and cover all parts of the instruction.\\
- Each segment must represent a single unit of instruction. If a step contains multiple actions (e.g., "add and divide"), split it into separate segments.\\
- Each segment must directly involve computation or logical derivation.\\
- Each segment only describes the action not the solution.\\
- The last segment must represent the final step such that the response to this segment would be the final answer.\\
- Do not include numeric or textual answers inside the segments. Only describe the action needed to obtain them.\\
- Limit the total number of segments to 6 or fewer.\\

Your output must be a list of these segments in the following JSON format:

[
  \{"segment": "[first step]"\},
  \{"segment": "[second step]"\},
  ...
]

Here are some examples:\\

Example Query:\\
Josh decides to try flipping a house. He buys a house for \$80k and then puts in \$50k in repairs. This increased the value of the house by 150\%. How much profit did he make? \\

Example Output:
[
  \{"segment": "Add the purchase price and repair costs to find total spending"\},
  \{"segment": "Calculate the increase in house value using 150\%"\},
  \{"segment": "Find the new value of the house after the increase"\},
  \{"segment": "Subtract total spending from the new value to get the profit"\}
]\\

Example Query:
It takes Sarah an average of 15 minutes and 36 seconds to solve 2 puzzles. If she wants to solve 12 puzzles at the same rate, it will take her X hours, Y minutes, and Z seconds. \\

Example Output:
[
  \{"segment": "find the time it takes to solve one puzzle"\},
  \{"segment": "Multiply the time for one puzzle by 12 to get the total time for 12 puzzles"\},
  \{"segment": "Convert the total seconds into hours, minutes, and seconds"\}
]\\

Now complete the task for the following question:\\
\{\{ question \}\}
  }}

  \end{tcolorbox}
\end{figure}
\newpage

\begin{figure}[H]
  \centering
  \begin{tcolorbox}[
    enhanced,
    fontupper=\small,
    colback=lightgray,
    colframe=darkgray,
    boxrule=0.8pt,
    arc=3pt,
    title={\small\textbf{Sub-Task Answer}},
    coltitle=white,
    colbacktitle=darkgray,
    boxsep=2pt,
    left=4pt,
    right=4pt,
    top=2pt,
    bottom=2pt,
    before upper={\setlength{\parskip}{1pt}},
    listing only,           
    listing options={       
        basicstyle=\ttfamily\small,
        breaklines=true
    }
]

  \colorbox{lightgray}{\parbox{\dimexpr\linewidth-2\fboxsep}{%

You will be given a question and step-by-step sequential segments that detail the reasoning or actions needed to answer the question. Your task is to solve each segment in order. For each segment, provide a clear step-by-step reasoning and the final result for that segment.\\

Your output must be a list of answers where each corresponds to each segment in the following JSON format:

[
  \{"explanation": "[detailed reasoning for the first segment]", "answer": "[final answer to the first segment]"\},
  \{"explanation": "[detailed reasoning for the second segment]", "answer": "[final answer to the second segment]"\},
  ...
]\\

Example:
Josh decides to try flipping a house. He buys a house for 80k and then puts in 50k in repairs. This increased the value of the house by 150\%. How much profit did he make? Format your answer as a JSON, like JSON = \{\"explanation\": <your step by step solution>, \"answer\": <final answer>\}.\\

[
  \{"segment": "Total amount Josh spent"\},
  \{"segment": "Increase in house value"\},
  \{"segment": "New value of the house"\},
  \{"segment": "Profit"\},
  \{"segment": "Format the final result as a JSON"\}
]\\

Output:
[
  \{"explanation": "Josh spent 80k to buy the house and 50k on repairs. Adding them gives 80k + 50k = 130k.", "answer": "130k"\},
  \{"explanation": "The original value was 80k. A 150\% increase means 1.5 × 80k = 120k.", "answer": "120k"\},
  \{"explanation": "The new value is original value + increase = 80k + 120k = 200k.", "answer": "200k"\},
  \{"explanation": "Profit = New value - Total spent = 200k - 130k = 70k.", "answer": "70k"\},
  \{"explanation": "Format the final result as a JSON.", "answer": "\{\"explanation\": \"Josh spent 80k to buy the house and 50k on repairs, totaling 130k. The house increased by 150\% of the original 80k, which is 120k. The new value of the house is 80k + 120k = 200k. Profit is new value minus total spent, 200k - 130k = 70k.\", \"answer\": \"70k\"\}"\}
]\\

Now complete the task for the following question:\\
\{question\}\\
\{sequential\_segments\}\\

Only output the JSON array, strictly following the format.\\
  }}

  \end{tcolorbox}
\end{figure}

\begin{figure}[H]
  \centering

  \begin{tcolorbox}[
    enhanced,
    fontupper=\small,
    colback=lightgray,
    colframe=darkgray,
    boxrule=0.8pt,
    arc=3pt,
    title={\small\textbf{Consistency}},
    coltitle=white,
    colbacktitle=darkgray,
    boxsep=2pt,
    left=4pt,
    right=4pt,
    top=2pt,
    bottom=2pt,
    before upper={\setlength{\parskip}{1pt}},
    listing only,           
    listing options={       
        basicstyle=\ttfamily\small,
        breaklines=true
    }
]
  \colorbox{lightgray}{\parbox{\dimexpr\linewidth-2\fboxsep}{%
  You are given a ground truth answer and a model answer.\\
Your task is to decide whether the two answers are equivalent in value.\\

**Rating Guidelines**\\
1. Ignore formatting differences (e.g., 2, "2", 2.0, answer: 2 should all be treated as the same).\\
2. Treat numbers written as words (e.g., two, forty-five) as equivalent to their numeric forms.\\
3. Units must be considered: 2 kg is not equal to 2 g, but 2000 g is equal to 2 kg.\\
4. Time expressions should be normalized: treat equivalent times as the same value even if expressed differently (e.g., "7:00", "7 am", "07:00", or "7 o’clock" all mean the same; "1 pm–3 pm" = "13:00–15:00"). Overlapping time intervals must match in value, regardless of format.\\
5. If either the model output or ground-truth answer is empty, missing, or unspecified, return a score of 0 regardless of other conditions.\\

**Model Output** \\
\{\{validator\_final\_answer\}\}\\

**Ground-truth Answer** \\
\{\{ground\_truth\}\}\\

**Scoring Criteria**\\
* Score 1: If the answers are equivalent in value.\\
* Score 0: If the answers are different in value.\\

Return your rating as a json of the form \{"score": your\_score, "justification": your\_justification\}. No additional explanation, text, or formatting outside the JSON.

  }}

  \end{tcolorbox}
\end{figure}

\begin{figure}[H]
  \centering
  \begin{tcolorbox}[
    enhanced,
    fontupper=\small,
    colback=lightgray,
    colframe=darkgray,
    boxrule=0.8pt,
    arc=3pt,
    title={\small\textbf{Scaffolded Variation}},
    coltitle=white,
    colbacktitle=darkgray,
    boxsep=2pt,
    left=4pt,
    right=4pt,
    top=2pt,
    bottom=2pt,
    before upper={\setlength{\parskip}{1pt}},
    listing only,           
    listing options={       
        basicstyle=\ttfamily\small,
        breaklines=true
    }
]
  \colorbox{lightgray}{\parbox{\dimexpr\linewidth-2\fboxsep}{%
  You will be given a question along with a partially completed step-by-step solution.\\
Your task is to rewrite the original question by incorporating the parts of the solution that have already been completed, so the rewritten question reflects that those steps are done. The new question should only require solving for the remaining steps. When rewriting, preserve the structure and wording of the original question as much as possible—only revise or replace the parts that are directly affected by the completed steps. Use the completed work to inform the new phrasing, as if someone is picking up the problem mid-way with that progress already understood. The rewritten question must be a single line without any line breaks.\\

Example\\
Original Question:\\
Josh decides to try flipping a house. He buys a house for \$80k and then puts in \$50k in repairs. This increased the value of the house by 150\%. How much profit did he make?\\

Solved Segments:\\
  \{"segment": "Calculate total amount Josh spent", "answer": "130K"\},\\
  \{"segment": "Calculate increase in house value", "answer": "120K"\},\\

Rewritten Question:\\
Josh decides to try flipping a house. He spent a total of \$130k for the house and the repairs. This increased the value of the house by \$120K. When the original value of the house was \$80k, how much profit did he make?\\

Now rewrite the following question using the completed steps provided. Reply in the following format:\\
Rewritten Question: <your rewritten version here>\\

Original Question:\\
\{\{ question \}\}\\

Solved Segments:\\
\{\{ solved\_sequential\_segments \}\}\\
  }}

  \end{tcolorbox}
\end{figure}

\begin{figure}[H]
  \centering
  \begin{tcolorbox}[
    enhanced,
    fontupper=\small,
    colback=lightgray,
    colframe=darkgray,
    boxrule=0.8pt,
    arc=3pt,
    title={\small\textbf{Skill Generation}},
    coltitle=white,
    colbacktitle=darkgray,
    boxsep=2pt,
    left=4pt,
    right=4pt,
    top=2pt,
    bottom=2pt,
    before upper={\setlength{\parskip}{1pt}},
    listing only,           
    listing options={       
        basicstyle=\ttfamily\small,
        breaklines=true
    }
]

  \colorbox{lightgray}{\parbox{\dimexpr\linewidth-2\fboxsep}{%
  You will be given a set of representative questions on \{\{category\}\}.

Your goal is to identify \{\{num\_of\_skills\}\} distinct skills that a learner (or model) must possess to correctly solve problems in this category. These skills will be used to map fine-grained sub-tasks to higher-level capabilities in order to analyze model performance.\\

Each skill should:\\
- Represent a specific cognitive or procedural competency, not a topic name (e.g., "Simplifying algebraic fractions" instead of "Rational expressions").\\
- Cover a range of difficulty, from foundational to advanced subskills.\\
- Be granular enough that multiple skills may be needed to solve a single question.\\
- Avoid redundancy (each skill should describe a unique capability).\\

Example:\\
Representative Questions (abridged):\\
1. Janet's ducks lay 16 eggs per day. She eats three for breakfast every morning and bakes muffins for her friends every day with four. She sells the remainder at the farmers' market daily for \$2 per fresh duck egg. How much in dollars does she make every day at the farmers' market? Steps to solve: "Determine the total number of eggs produced each day", "Subtract the three eggs Janet eats for breakfast from the total", "Subtract the four eggs used for muffins from the remaining eggs", "Calculate the number of eggs left after breakfast and muffins, which are sold", "Multiply the number of eggs sold by \$2 to find the daily earnings at the farmers' market."\\
2. Henry and 3 of his friends order 7 pizzas for lunch. Each pizza is cut into 8 slices. If Henry and his friends want to share the pizzas equally, how many slices can each of them have? Steps to solve: "Multiply the number of pizzas (7) by the number of slices per pizza (8) to find the total number of slices", "Count the total number of people sharing the pizza (Henry plus 3 friends) to determine the number of recipients", "Divide the total number of slices by the number of people to find how many slices each person receives."\\
3. Jame will turn 27 in 5 years. In 8 years his cousin will be 5 years younger than twice his age. How many years separate the age of the two now? Steps to solve: "Subtract 5 from 27 to obtain Jame's current age", "Add 8 to Jame's current age to find Jame's age in 8 years", "Formulate cousin's age in 8 years as (twice Jame's age in 8 years) minus 5", "Calculate cousin's age in 8 years using the value from step 2", "Subtract 8 from cousin's age in 8 years to get cousin's current age", "Compute the absolute difference between cousin's current age and Jame's current age."\\

Example Skills (abridged):\\
1. Quantitative interpretation of word problems: extracting relevant numbers, entities, and constraints from text.\\
2. Basic arithmetic operations: addition, subtraction, multiplication, and division in multi-step contexts.\\
3. Sequential quantity tracking: updating remaining or accumulated values across successive steps.\\
4. Equal sharing and unit-rate calculations: dividing quantities evenly and applying per-unit values.\\
5. Temporal and relational reasoning: shifting values across time and forming linear relationships from verbal descriptions.\\

Now perform this task:\\
Using the representative questions below, infer and list \{\{num\_of\_skills\}\} well-defined skills required to solve the full set of problems.\\

Focus on the underlying reasoning and computation abilities shared across questions, rather than restating the solution steps or question content.\\

Representative Questions:\\
\{\{question\_clusters\}\}\\

Output Format:\\
- Return a numbered list of \{\{num\_of\_skills\}\} skills, each described in one concise sentence.\\
- Do not include solutions, examples, or references to specific questions.\\
- Each item should be in the format: 1. Skill name: brief explanation.\\

  }}

  \end{tcolorbox}
\end{figure}

\begin{figure}[H]
  \centering
  \begin{tcolorbox}[
    enhanced,
    fontupper=\small,
    colback=lightgray,
    colframe=darkgray,
    boxrule=0.8pt,
    arc=3pt,
    title={\small\textbf{Skill Mapping}},
    coltitle=white,
    colbacktitle=darkgray,
    boxsep=2pt,
    left=4pt,
    right=4pt,
    top=2pt,
    bottom=2pt,
    before upper={\setlength{\parskip}{1pt}},
    listing only,           
    listing options={       
        basicstyle=\ttfamily\small,
        breaklines=true
    }
]

  \colorbox{lightgray}{\parbox{\dimexpr\linewidth-2\fboxsep}{%
  You are a classification model.\\
You will be given a problem and its associated sub-tasks. 
For each sub-task, classify it into exactly one of the predefined skills listed below. \\

Skills (skill ID: description): \\
\{\{skills\}\}\\

Output requirements:\\
- Return a JSON array with one object per sub-task.\\
- Each object must contain only the selected skill ID from the list.\\
- Do not include any extra text, explanation, or formatting.\\

Example format:
[ 
  \{"skill": skill id from the list\},
  \{"skill": skill id from the list\},
  \{"skill": skill id from the list\},
  ...
]\\

Problem:\\ 
\{\{ question \}\}

Sub-Tasks:\\ 
\{\{ sub-tasks \}\}
  }}

  \end{tcolorbox}
\end{figure}

\newpage

\newpage
\section{Results}

\begin{table}[H]
\centering
\caption{Individual skill accuracy across models and datasets (Top-15 most frequent skills). The two lowest-performing skills are highlighted in bold.}
\small
\begin{tabular}{l|cccccc}
\hline
\textbf{Skill (ToT Arithmetic)} & \textbf{Qwen3B} & \textbf{Qwen7B} & \textbf{Llama3B} & \textbf{Llama8B} & \textbf{Granite2B} & \textbf{Granite8B} \\ \hline
Structured JSON output & {0.55} & {0.70} & {0.38} & {0.42} & {0.53} & {0.66} \\
Arithmetic on durations & 0.54 & 0.71 & 0.42 & 0.50 & 0.55 & 0.69 \\
Unit conversion for time spans & 0.51 & 0.70 & 0.33 & 0.42 & 0.37 & 0.60 \\
Calendar arithmetic  &{0.18} & 0.22 & 0.19 & \textbf{0.18} & 0.16 & 0.23 \\
Natural-language date/time parsing  & 0.35 & 0.27 & 0.23 & 0.26 & 0.28 & 0.27 \\
Discrete slot counting  & 0.19 & 0.29 & 0.15 & 0.21 & 0.16 & 0.25 \\
Overlap and intersection of time  & \textbf{0.11} & \textbf{0.18} & \textbf{0.13} & 0.20 & \textbf{0.11} & \textbf{0.22} \\
Multi-format date conversion  & 0.62 & 0.65 & 0.42 & 0.52 & 0.55 & 0.64 \\
Day-of-week determination  & 0.44 & 0.37 & 0.27 & 0.36 & 0.28 & 0.43 \\
Chronological ordering of events  & \textbf{0.02} & \textbf{0.04} & \textbf{0.08} & \textbf{0.14} & \textbf{0.08} & \textbf{0.10} \\
Extremum identification  & 0.26 & 0.19 & 0.30 & 0.32 & 0.26 & 0.31 \\
Normalization of incomplete timestamps  &  0.39 & 0.47 & 0.63 & 0.60 & 0.37 & 0.49 \\
Midnight and date-boundary wrap-around  & 0.28 & 0.35 & 0.19 & 0.28 & 0.30 & \textbf{0.22} \\
Time-zone conversion  & 0.40 & 0.62 & 0.55 & 0.66 & 0.36 & 0.57 \\
Relative-date reasoning  & 0.29 & 0.44 & 0.20 & 0.31 & 0.12 & 0.24 \\
\textbf{Average (std)} & 0.34\,(0.17) & 0.41\,(0.22) & 0.30\,(0.16) & 0.36\,(0.16) & 0.30\,(0.16) & 0.39\,(0.20) \\
\hline
\end{tabular}

\vspace{2mm}

\begin{tabular}{l|cccccc}
\hline
\textbf{Skill (GSM8K)} & \textbf{Qwen3B} & \textbf{Qwen7B} & \textbf{Llama3B} & \textbf{Llama8B} & \textbf{Granite2B} & \textbf{Granite8B} \\ \hline
Sequential quantity tracking & 0.70 & 0.77 & 0.74 & 0.70 & 0.66 & 0.77 \\
Using unit rates to compute totals & 0.82 & 0.82 & 0.86 & 0.87 & 0.75 & 0.83 \\
Extracting quantitative data from text & \textbf{0.25} & \textbf{0.19} & \textbf{0.29} & \textbf{0.20} & \textbf{0.11} & \textbf{0.19} \\
Summing contributions from multi sources & 0.67 & 0.78 & 0.77 & 0.80 & 0.69 & 0.89 \\
Formulating and solving linear equations & \textbf{0.49} & \textbf{0.50} & \textbf{0.42} & \textbf{0.48} & 0.51 & \textbf{0.42} \\
Multi-digit multiplication and division & 0.90 & 0.85 & 0.84 & 0.94 & 0.73 & 0.83 \\
Proportional reasoning & 0.77 & 0.76 & 0.77 & 0.85 & 0.57 & 0.79 \\
Calculating a percentage of a quantity &  0.85 & 0.89 & 0.93 & 0.96 & 0.67 & 0.93 \\
Performing operations with fractions & 0.70 & 0.80 & 0.73 & 0.69 & 0.60 & 0.58 \\
Multiplicative scaling for unknowns & 0.81 & 0.86 & 0.76 & 0.73 & 0.74 & 0.75 \\
Remaining amount after purchases & 0.76 & 0.89 & 0.86 & 0.84 & 0.83 & 0.82 \\
Total cost from price and quantity & 0.88 & 0.91 & 0.79 & 0.80 & 0.67 & 1.00 \\
Isolate unknown quantities & 0.86 & 1.00 & 0.76 & 0.73 & 0.68 & 0.71 \\
Equal sharing of a total & 0.89 & 1.00 & 0.74 & 0.79 & 0.59 & 0.83 \\
Determining remaining quantity & 0.82 & 1.00 & 0.90 & 0.80 & \textbf{0.50} & 1.00 \\
\textbf{Average (std)}& 0.74\,(0.17)& 0.80\,(0.21)& 0.74\,(0.17)& 0.75\,(0.19)& 0.62\,(0.17)& 0.76\,(0.22) \\
\hline
\end{tabular}

\vspace{2mm}

\begin{tabular}{l|cccccc}
\hline
\textbf{Skill (Math-Hard)} & \textbf{Qwen3B} & \textbf{Qwen7B} & \textbf{Llama3B} & \textbf{Llama8B} & \textbf{Granite2B} & \textbf{Granite8B} \\ \hline
Multi-step integer/fraction arithmetic & 0.45 & 0.51 & 0.55 & 0.60 & 0.56 & 0.63 \\
Translating problems into algebra & 0.24 & 0.23 & 0.31 & \textbf{0.31} & \textbf{0.34} & 0.34 \\
Solving linear equations and systems & 0.29 & 0.31 & 0.41 & 0.45 & 0.46 & 0.47 \\
Combinatorial counting & 0.36 & 0.46 & 0.44 & 0.42 & 0.43 & 0.47 \\
Quadratic equation techniques & 0.30 & 0.40 & 0.44 & 0.51 & 0.50 & 0.42 \\
Modular arithmetic and congruence & 0.34 & 0.44 & 0.40 & 0.35 & 0.39 & 0.54 \\
GCD/LCM via prime factorization & 0.38 & 0.41 & 0.49 & 0.51 & 0.49 & 0.53 \\
Probability via counting & 0.28 & 0.46 & 0.39 & 0.39 & 0.44 & 0.49 \\
Domain and range of functions & 0.20 & 0.20 & 0.55 & 0.49 & 0.43 & 0.50 \\
Arithmetic and geometric series analysis & 0.36 & 0.28 & 0.43 & 0.41 & 0.52 & 0.35 \\
Completing the square for optimization & 0.20 & \textbf{0.12} & 0.68 & 0.46 & 0.46 & 0.62 \\
Calculus-based optimization & 0.23 & 0.35 & 0.39 & \textbf{0.27} & 0.35 & 0.36 \\
Absolute-value inequality manipulation & \textbf{0.17} & 0.30 & 0.35 & 0.38 & 0.45 & \textbf{0.17} \\
Symmetry counting w/ Burnside's Lemma & 0.28 & \textbf{0.19} & \textbf{0.22} & 0.35 & \textbf{0.18} & 0.33 \\
Solving higher-degree polynomials & \textbf{0.09} & 0.24 & \textbf{0.30} & 0.35 & 0.39 & \textbf{0.30} \\
\textbf{Average (std)}& 0.28\,(0.09)& 0.33\,(0.12)& 0.42\,(0.11)& 0.42\,(0.09)& 0.43\,(0.09)& 0.43\,(0.13) \\

\hline
\end{tabular}

\label{tab:method1_skill_accuracy_per_dataset_copy}
\end{table}

\begin{table}[H]
\centering
\small
\caption{Mean minimum scaffolding level ($k$) per skill combination. $k=0$: solved independently; \%\,$k>0$: solvable with scaffolding; \% Intractable: unsolvable even with full scaffolding. Higher $k$ indicates greater difficulty.}
\label{tab:scaffolding_copy}
\begin{tabular}{lcccccc}
\hline
\textbf{Combination (ToT)} & \textbf{\#Tasks} & \textbf{Model} & \textbf{Mean $k$} & \textbf{\% $k=0$} & \textbf{\% $k>0$} & \textbf{\% Intractable} \\
\hline
\multirow{6}{*}{\shortstack[l]{Overlapping + Overlapping + \\
Meeting slots + Meeting slots + \\
JSON}}
& \multirow{6}{*}{65} & Qwen3B    & 3.02 & 6.15 & 70.76 & 23.07 \\
&                     & Qwen7B    & 2.51 & 6.15 & 89.23 & 4.61 \\
&                     & Llama3B   & 3.06 & 3.07 & 75.38 & 21.53 \\
&                     & Llama8B   & 2.69 & 10.76 & 81.53 & 7.69 \\
&                     & Granite2B & 2.93 & 1.53 & 73.84 & 24.61 \\
&                     & Granite8B & 2.79 & 10.76 & 81.53 & 7.69 \\
\hline
\multirow{6}{*}{\shortstack[l]{Unit + Duration + Duration + \\
Unit + JSON}}
& \multirow{6}{*}{61} & Qwen3B    & 1.90 & 45.90 & 52.45 & 1.63 \\
&                     & Qwen7B    & 1.45 & 67.21 & 32.78 & 0.0 \\
&                     & Llama3B   & 3.68 & 9.83 & 31.14 & 59.01 \\
&                     & Llama8B   & 3.44 & 13.11 & 59.01 & 27.86 \\
&                     & Granite2B & 3.19 & 9.83 & 59.01 & 31.14 \\
&                     & Granite8B & 1.97 & 21.31 & 78.68 & 0.0 \\
\hline
\multirow{6}{*}{\shortstack[l]{Unit + Date-Boundary + \\
Duration + Unit + JSON}}
& \multirow{6}{*}{41} & Qwen3B    & 2.32 & 4.87 & 90.24 & 4.87 \\
&                     & Qwen7B    & 2.18 & 34.14 & 65.85 & 0.00 \\
&                     & Llama3B   & 3.85 & 0.0 & 68.29 & 31.70 \\
&                     & Llama8B   & 3.60 & 0.0 & 80.48 & 19.51 \\
&                     & Granite2B & 3.82 & 2.43 & 68.29 & 29.26 \\
&                     & Granite8B & 2.48 & 9.75 & 85.36 & 4.87 \\
\hline
\end{tabular}

\vspace{2mm}

\begin{tabular}{lcccccc}
\hline
\textbf{Combination (GMS8k)} & \textbf{\#Tasks} & \textbf{Model} & \textbf{Mean $k$} & \textbf{\% $k=0$} & \textbf{\% $k>0$} & \textbf{\% Intractable} \\
\hline
\multirow{6}{*}{\shortstack[l]{Unit-Total + Unit-Total + \\
Unit-Total + Sum }}
& \multirow{6}{*}{15} & Qwen3B    & 0.00 & 93.33 & 0.00 & 6.66 \\
&                     & Qwen7B    & 0.00 & 100.00 & 0.00 & 0.00 \\
&                     & Llama3B   & 1.33 & 80.00 & 20.00 & 0.00 \\
&                     & Llama8B   & 1.00 & 93.33 & 6.66 & 0.00 \\
&                     & Granite2B & 1.00 & 93.33 & 6.66 & 0.00 \\
&                     & Granite8B & 2.00 & 93.33 & 6.66 & 0.00 \\
\hline
\multirow{6}{*}{\shortstack[l]{Quantity-Tracking + \\
Quantity-Tracking}}
& \multirow{6}{*}{14} & Qwen3B    & 1.00 & 92.85 & 7.14 & 0.00 \\
&                     & Qwen7B    & 0.00 & 100.00 & 0.00 & 0.00 \\
&                     & Llama3B   & 1.00 & 85.71 & 14.28 & 0.00 \\
&                     & Llama8B   & 0.00 & 100.00 & 0.00 & 0.00 \\
&                     & Granite2B & 1.00 & 92.85 & 7.14 & 0.00 \\
&                     & Granite8B & 1.00 & 92.85 & 7.14 & 0.00 \\
\hline
\multirow{6}{*}{\shortstack[l]{Extracting + Proportional + \\
Quantity-Tracking}}
& \multirow{6}{*}{10} & Qwen3B    & 0.00 & 100.00 & 0.00 & 0.00 \\
&                     & Qwen7B    & 0.00 & 100.00 & 0.00 & 0.00 \\
&                     & Llama3B   & 0.00 & 100.00 & 0.00 & 0.00 \\
&                     & Llama8B   & 0.00 & 100.00 & 0.00 & 0.00 \\
&                     & Granite2B & 0.00 & 100.00 & 0.00 & 0.00 \\
&                     & Granite8B & 1.00 & 90.00 & 10.00 & 0.00 \\
\hline
\end{tabular}

\vspace{2mm}

\begin{tabular}{lcccccc}
\hline
\textbf{Combination (Math-Hard)} & \textbf{\#Tasks} & \textbf{Model} & \textbf{Mean $k$} & \textbf{\% $k=0$} & \textbf{\% $k>0$} & \textbf{\% Intractable} \\
\hline
\multirow{6}{*}{\shortstack[l]{Arithmetic + Arithmetic + \\
Arithmetic + Arithmetic + \\
Arithmetic}}
& \multirow{6}{*}{12} & Qwen3B    & 2.14 & 33.33 & 58.33 & 8.33 \\
&                     & Qwen7B    & 2.5 & 75.0 & 16.66 & 8.33 \\
&                     & Llama3B   & 1.71 & 33.33 & 58.33 & 8.33 \\
&                     & Llama8B   & 1.00 & 41.66 & 41.66 & 16.66 \\
&                     & Granite2B & 2.50  & 41.66 & 50.00 & 8.33 \\
&                     & Granite8B & 2.00 & 75.00 & 25.00 & 0.00 \\
\hline
\multirow{6}{*}{\shortstack[l]{Combinatorics + Combinatorics + \\
Combinatorics + Combinatorics + \\
Combinatorics}}
& \multirow{6}{*}{11} & Qwen3B    & 1.8 & 36.36 & 45.45 & 18.18 \\
&                     & Qwen7B    & 2.42 & 27.27 & 63.63 & 9.09 \\
&                     & Llama3B   & 2.62 & 18.18 & 72.72 & 9.09 \\
&                     & Llama8B   & 2.16 & 18.18 & 54.54 & 27.27 \\
&                     & Granite2B & 1.88 & 9.09 & 81.81 & 9.09 \\
&                     & Granite8B & 1.75 & 63.63 & 36.36 & 0.00 \\
\hline
\multirow{6}{*}{\shortstack[l]{Congruences + Congruences + \\
Congruences + Congruences + \\
Congruences + Congruences}}
& \multirow{6}{*}{9} & Qwen3B    & 2.88 & 0.00 & 100.00 & 0.00 \\
&                     & Qwen7B    & 2.50 & 11.11 & 88.88 & 0.00 \\
&                     & Llama3B   & 3.14 & 22.22 & 77.77 & 0.00 \\
&                     & Llama8B   & 3.00 & 11.11 & 77.77 & 11.11 \\
&                     & Granite2B & 3.00 & 0.00 & 88.88 & 11.11 \\
&                     & Granite8B & 3.00 & 22.22 & 77.77 & 0.00 \\
\hline
\end{tabular}

\end{table}

\begin{table*}[p]
\centering
\small
\caption{Frequency of compositional-skill bottlenecks. Numbers indicate count of scaffolded cases. Only the first word of each skill is shown.}
\label{tab:skill_bottlenecks_top10}
\begin{tabular}{llllllll}
\hline
              & \textbf{Skill (ToT)}      & \textbf{Qwen3B} & \textbf{Qwen7B} & \textbf{Llama3B} & \textbf{Llama8B} & \textbf{Granite2B} & \textbf{Granite8B} \\
\hline
\multirow{30}{*}{\rotatebox{90}{\shortstack[l]{$n=1$}}} 
& \textbf{ToT Arithmetic}  &  &  &  &  &  &  \\
& Unit       & \textbf{45} & \textbf{48} & 10 & 19 & 16 & 49 \\
& Natural    & \textbf{43} & \textbf{52} & \textbf{42} & \textbf{44} & 30 & \textbf{59} \\
& Overlap    & 27 & 39 & 22 & 28 & 20 & 28 \\
& Calendar   & 26 & 40 & 28 & 38 & 34 & 36 \\
& Day-of-week & 25 & 5 & 20 & 16 & 21 & 17 \\
& Arithmetic & 18 & 20 & 3  & 12 & 7  & 28 \\
& Time-zone  & 15 & 6  & 13 & 11 & 15 & 15 \\
& Relative   & 10 & 6  & 2  & 12 & 5  & 7  \\
& Normalization & 7 & 12 & 6 & 8 & 6 & 6 \\
& Multi-format & 4 & 7 & 5 & 6 & 5 & 4 \\ 
& \textbf{GSM8K}  &  &  &  &  &  &  \\
& Unit       & 33 & 7  & 35 & 19 & 24 & 13 \\
& Extracting & 22 & 11 & 14 & 13 & 18 & 10 \\
& Sequential & 16 & 7  & 21 & 13 & 11 & 12 \\
& Multi-digit & 9  & 4  & 14 & 10 & 10 & 4 \\
& Performing & 8  & -  & 10 & 3  & 4  & 3 \\
& Proportional & 8 & 3  & 6  & 5  & 6  & 5 \\
& Calculating & 7 & 2  & 10 & 7  & 5  & 3 \\
& Total & 6 & 3 & 6 & 2 & 6 & - \\
& Formulating & 4 & - & 8 & 2 & 3 & 6 \\
& Summing & 3 & - & - & 2 & - & - \\ 
& \textbf{Math-Hard}  &  &  &  &  &  &  \\
& Translating       & 41 & 43 & 43 & 50 & 50 & 63 \\
& Multi-step        & 26 & 17 & 29 & 20 & 20 & 18 \\
& Combinatorial     & 20 & 28 & 29 & 25 & 27 & 24 \\
& Modular           & 16 & 13 & 11 & 11 & 16 & 11 \\
& Probability       & 9  & 7  & 6  & 7  & 11 & 6  \\
& Solving           & 8  & 6  & 3  & 6  & 7  & 0  \\
& Quadratic         & 4  & 4  & 3  & 4  & 3  & 0  \\
& GCD/LCM           & 4  & 5  & 4  & 4  & 3  & 5  \\
& Arithmetic        & 3  & 4  & 3  & 4  & 9  & 5  \\
& Domain            & 3  & 6  & 3  & 6  & 5  & 4  \\
\hline
\multirow{27}{*}{\rotatebox{90}{\shortstack[l]{$n=2$}}} 
& \textbf{ToT Arithmetic} &&&&& \\
& Overlap / Discrete & \textbf{56} & \textbf{49} & \textbf{58} & \textbf{59} & \textbf{66} & \textbf{59} \\
& Natural / Calendar  & 39 & 30 & 34 & 41 & \textbf{43} & 50 \\
& Calendar / Arithmetic & 31 & 25 & 21 & 25 & 34 & 35 \\
& Unit / Arithmetic  & 30 & 17 & 36 & \textbf{57} & \textbf{54} & \textbf{53} \\
& Multi / Calendar   & 23 & 15 & 15 & 17 & 19 & 16 \\
& Relative / Calendar & 15 & 14 & 12 & 13 & 20 & 21 \\
& Natural / Arithmetic & 12 & 4 & 13 & 13 & 9 & 11 \\
& Unit / Midnight    & 12 & 9 & 10 & 4 & 12 & 13 \\
& Natural / Overlap  & 9  & 25 & 6  & 12 & 8  & 11 \\
& Unit / Calendar    & 7  & 9  & 11 & 8  & 10 & 13 \\ 
& \textbf{GSM8K} &&&&& \\
& Extracting / Sequential & 5 & 2 & 5 & 2 & 7 & 3 \\
& Unit / Summing & 3 & - & - & - & 2 & - \\
& Extracting / Multiplicative & 3 & - & 3 & - & - & - \\
& Extracting / Proportional & 2 & - & 2 & 3 & 2 & - \\
& Unit / Proportional & 2 & - & 2 & - & 2 & - \\
& Unit / Sequential & 2 & - & - & - & - & - \\
& Sequential / Proportional & 2 & 2 & - & - & 2 & 2 \\ 
& \textbf{Math-Hard}  &  &  &  &  &  &  \\
& Translating / Solving       & 15 & 10 & 11 & 10 & 13 & 4 \\
& Translating / Multi-step    & 6  & 6  & 5  & 10 & 10 & 7 \\
& Domain / Quadratic          & 2  & 2  & 0  & 2  & 0  & 0 \\
& Arithmetic / Translating    & 2  & 2  & 3  & 3  & 3  & 3 \\
& Combinatorial / Translating & 2  & 2  & 2  & 4  & 5  & 2 \\
& Modular / Multi-step        & 0  & 2  & 2  & 0  & 0  & 0 \\
& Probability / Combinatorial & 2  & 2  & 3  & 3  & 2  & 2 \\
& Translating / Quadratic     & 0  & 4  & 4  & 4  & 5  & 0 \\
& Multi-step / GCD/LCM        & 0  & 2  & 0  & 0  & 0  & 0 \\
& Calculus / Translating      & 0  & 0  & 0  & 0  & 3  & 2 \\

\hline
\multirow{12}{*}{\rotatebox{90}{\shortstack[l]{$n>=3$}}}
& \textbf{ToT Arithmetic} &&&&& \\
& Natural / Overlap / Discrete & 29 & 29 & \textbf{38} & 26 & 42 & 32 \\
& Unit / Midnight / Arithmetic & 14 & 11 & 27 & 29 & 26 & 18 \\
& Natural / Calendar / Multi   & 5  & 2  & 4  & 5  & 4  & 7  \\
& Unit / Arithmetic / Midnight & 4  & 3  & 5  & 7  & 3  & 4  \\
& Relative / Calendar / Multi  & 4  & 2  & 9  & 3  & 4  & 3  \\
& Natural / Arithmetic / Day-of-week & 4 & 4 & 5 & 4 & 0 & 4 \\
& Unit / Calendar / Multi      & 3  & 2  & 2  & 4 & 3  & 2  \\
& Arithmetic / Unit / Midnight & 3  & 3  & 3  & 7 & 3  & 4  \\
& Unit / Time-zone / Arithmetic / Midnight & 2 & 3 & 2 & 3 & 2 & 3 \\
& Natural / Calendar / Unit / Arithmetic & 2 & 2 & 2 & 2 & 2 & 2 \\
& \textbf{Math-Hard}  &  &  &  &  &  &  \\
& Translating / Multi-step / Solving & 0 & 2 & 2 & 2 & 3 & 2 \\
\hline
\end{tabular}
\end{table*}

\end{document}